\documentclass{article}

\usepackage[preprint]{neurips_2025}

\usepackage[utf8]{inputenc} 
\usepackage[T1]{fontenc}    
\usepackage{hyperref}       
\usepackage{url}            
\usepackage{booktabs}       
\usepackage{amsfonts}       
\usepackage{nicefrac}       
\usepackage{microtype}      
\usepackage{xcolor}         

\usepackage{graphicx}
\usepackage{wrapfig}
\usepackage{subcaption}
\usepackage{amsmath}
\usepackage{bm}
\usepackage{multirow}
\usepackage{natbib}
\usepackage[dvipsnames]{xcolor}

\newtheorem{lemma}{Lemma}  

\title{Verifying Large Language Models' Reasoning Paths via Correlation Matrix Rank}

\author{%
Jiayu Liu\textsuperscript{1,2} \quad Wei Dai\textsuperscript{1} \quad Zhenya Huang\textsuperscript{2,3}\thanks{Corresponding Authors} \quad Ning Miao\textsuperscript{4,5} \quad Enhong Chen\textsuperscript{2,3}\\
1: School of Artificial Intelligence and Data Science, University of Science \\ and Technology of China \\
2: State Key Laboratory of Cognitive Intelligence \\
3: School of Computer Science and Technology, University of Science \\ and Technology of China \\
4: Department of Data Science, City University of Hong Kong \\
5: Hong Kong Institute of AI for Science, City University of Hong Kong \\
\texttt{jy251198@mail.ustc.edu.cn}
}

\begin{document}

\maketitle

\begin{abstract}
Despite the strong reasoning ability of large language models~(LLMs), they are prone to errors and hallucinations. As a result, how to check their outputs effectively and efficiently has become a critical problem in their applications. Existing checking methods heavily rely on external resources, such as trained verifiers (e.g., process/outcome reward models) or elaborate prompts, which lead to high computational overhead and are only applicable to specific domains. In this paper, we investigate whether the internal behaviors of LLMs have already implied the credibility of their reasoning paths. Specifically, we find that the rank of the correlation matrix between the input problem and the output reasoning path is a robust indicator of reasoning correctness. Different from other correctness indicators for LLMs, the calculation of the correlation matrix only relies on the LLM itself, which avoids the hassle of training a separate model or designing complicated prompts. Based on it, we design a simple, plug-and-play \emph{Self-Indicator} method to reweight candidate reasoning paths, which achieves significant performance improvements than other voting and verification methods with very few computational overhead. Our experiments across multiple LLMs of varying scales and model families have further shown the effectiveness of Self-Indicator. It achieves over 75\% accuracy in distinguishing correct reasoning paths from incorrect ones, and, in turn, improves the accuracies on three reasoning benchmarks by more than 8\%.
\end{abstract}

\section{Introduction}
Despite demonstrating impressive performances in several reasoning tasks, large language models (LLMs) still remain prone to errors and hallucinations~\cite{chang2024survey}. To address it, a critical area of current research focuses on how to quickly and automatically verify the correctness of their outputs, which can be utilized to select candidate outputs or provide feedback for the reflection of LLMs~\cite{pan2023automatically,weng2023large,shinn2023reflexion,goucritic}. 
Along this line, existing studies have tried to retrieve knowledge~\cite{yu2023improving,peng2023check,huo2023retrieving}, leverage additional (trained) models or tools~\cite{xue2024decompose,cobbe2021training}, or prompt LLMs to check their own outputs directly~\cite{miaoselfcheck,weng2023large,ling2023deductive}. While they have shown promise, the reliance on \emph{external} sources may restrict their scalability and applicability in broader settings, while lead to significant computational overheads.

Then, it is natural to ask whether we can check the reasoning paths of an LLM fully by itself, or more fundamentally, \emph{Does an LLM inherently ``know'' when it is correct?} 
More specifically, we would like to know whether there are any patterns in the final or the intermediate outputs of an LLM that indicates the correctness of its reasoning path.
Previous studies have partially answered this question. For example, \cite{chen2024context,geva2023dissecting,halawioverthinking} examined the task knowledge-based question answering~(QA) and found that when predicting an object, the context activation for correct entities tends to be sharper than for incorrect one. 
Similarly, \cite{yuksekgonulattention} found that the attention to specific tokens correlates with LLMs' correctness in factual queries. 
However, these works primarily focus on the correctness of single-token output in factual QA and are unsuitable to assess the correctness of complex reasoning paths, which are typically longer and may contain multiple steps.

\begin{wrapfigure}{r}{0.6\linewidth}
  \centering
  \vspace{-10pt}
  \setlength{\abovecaptionskip}{0pt}
  \begin{subfigure}[b]{0.6\textwidth}
    \centering
    \includegraphics[width=\textwidth]{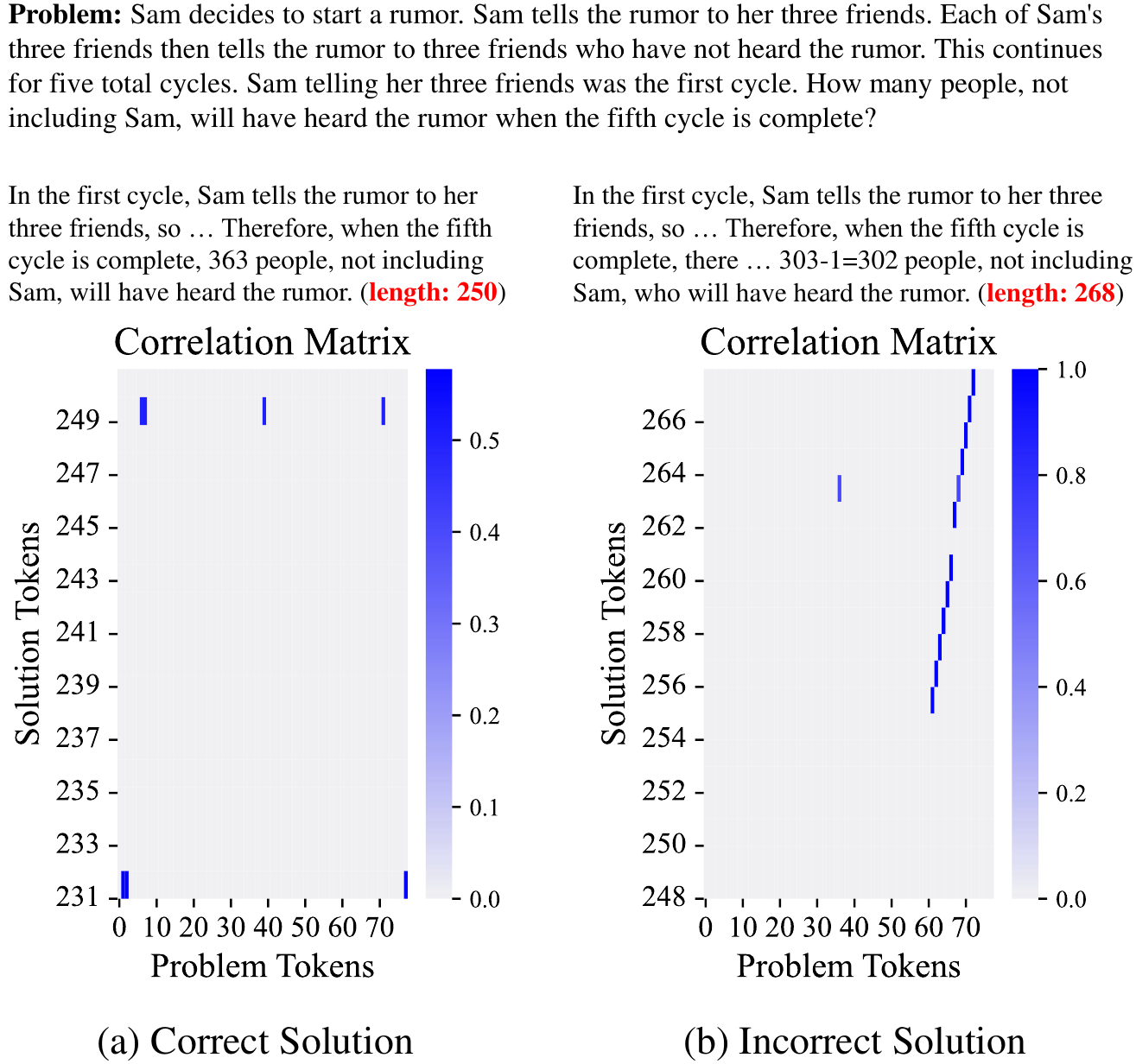}
  \end{subfigure}
  \hfill
  \caption{Correlation matrices (detailed in Section~\ref{section:motivation}) of two solutions  generated by LLaMA2-13B. Since the first halves of the two solutions are highly similar and correct, we present only the matrices corresponding to the final 20 tokens for clarity.}
  \label{fig:preliminary}
  \vspace{-8pt}
\end{wrapfigure}
In this paper, we extend the analysis to evaluating the correctness of reasoning paths. We find that, there are usually a few key patterns in the input problem. While correct solutions mainly focus on the key patterns, incorrect ones tend to be distracted by other spurious patterns. 
As a result, \emph{the correlation between problem and correct solutions has lower complexity than incorrect solutions}. 
In practice, we use the rank of correlation matrices between problem and solution token representations as a proxy for the correlation complexity, which we call \textbf{Self-Indicator}. It is convenient to use without requiring any external resources.
Figure~\ref{fig:preliminary} shows two fragments of the correlation matrices between correct/incorrect reasoning paths and the problem. 
We can easily see that the correlation matrix between the incorrect solution and the problem has significantly higher rank. 
One major reason is that the final reasoning step in the solution exhibits spurious correlations with the inquiry part of the problem (top-right corner), which indicates that the model is over-attending to tokens such as ``not including Sam'' and performs an unrelated and wrong subtraction ``303-1=302''. In contrast, the final tokens in the correct solution focus on summarizing previous steps, thus exhibiting much sparser correlations with the original problem tokens.
Based on this insight, we apply our Self-Indicator to improve reasoning reliability by designing a policy that generates multiple reasoning paths and then ranks and weights the candidate solutions using our Self-Indicator score.

To verify the effectiveness of our method in recognizing incorrect reasoning paths and, in turn, improving LLMs' reasoning performance, we conduct experiments on three different LLMs, including LLaMA2-13B~\citep{touvron2023llama}, LLaMA3-70B~\citep{meta2024introducing}, and GPT-3.5-Turbo~\citep{chatgpt2023}. 
We find that Self-Indicator achieves over 75\% accuracy in distinguishing correct from incorrect solutions. 
Interestingly, we find that we can generate reasoning paths from one LLM and calculate correlation matrices with another LLM, which makes Self-Indicator an efficient verification method for large or proprietary models.
Moreover, we observe that using Self-Indicator to reweight candidate reasoning paths can bring significant performance gains on three mathematical reasoning benchmarks, with very few overhead in computation compared to other voting or verification baselines.

Our contributions can be summarized as follows:
\begin{itemize}
  \item We propose a novel self-assessment metric, \emph{Self-Indicator}, which can efficiently evaluate the correctness of LLM-generated reasoning paths without relying on any external resources. We theoretically and empirically verify its reliability across multiple LLMs.
  
  \item We demonstrate that \emph{Self-Indicator} provides a lightweight and plug-and-play approach to enhance LLMs' reasoning ability by reweighting candidate reasoning paths.

  \item We validate our method with three representative LLM backbones, improving accuracy by over 8\% and validating its effectiveness, generality, and cost-efficiency.
\end{itemize}

\section{Related Work}
We categorize current research on automatic LLM assessment into three directions. 

\textbf{Representation-based.} This line of research explores whether the model's internal representations inherently encode correctness signals. Prior studies suggest that language models store structured factual knowledge (e.g., subject-relation-object triples) in the parameters~\cite{petroni2019language,geva2021transformer}, and their activations or attention patterns exhibit distinct behaviors when recalling correct versus incorrect facts~\cite{meng2022locating,geva2023dissecting}. Following this intuition,~\cite{yuksekgonulattention} reveals the positive relationship between an LLM's correctness and its attention to constraint tokens during generation. On this basis, they train a regression model to estimate the probability of factual constraint satisfaction using attention weights across all model layers. ~\cite{chen2024context} identifies a pattern where context activations become less sharp when hallucinations occur. By incorporating a corresponding entropy-based metric into the decoding process, they allow the model to dynamically adjust its outputs based on activation sharpness. However, these studies mainly focus on evaluating the correctness of entities in knowledge triples, which are hard to generalize to general reasoning tasks where solutions are represented by rationales.

\textbf{Prompt engineering-based.} The second direction leverages prompt engineering to elicit self-assessment capabilities of LLMs. For example, Self-Verification~\cite{weng2023large} leverages the LLM to perform backward verification by assessing whether the generated answer remains consistent with the given conditions. Deductive Verification~\cite{ling2023deductive} decomposes a reasoning verification process into a series of subprocesses and introduces a deductive reasoning format called Natural Program to enable explicit and step-by-step self-verification. SelfCheck~\cite{miaoselfcheck} decomposes the verification process into four stages, requiring the LLM to sequentially extract the target, collect relevant information, regenerate the reasoning step, and compare with the original results. This structured approach ensures a more rigorous and fine-grained self-assessment mechanism. However, these approaches often require additional LLM calls and are heavily constrained by the model's own evaluation capabilities~\cite{huanglarge}.

\textbf{External sources-based.} This direction relies on external resources, such as knowledge bases, tools, or specially trained verifiers, to evaluate LLMs' outputs. Knowledge bases are primarily used in knowledge-intensive tasks, where they help verify if LLM responses are consistent with existing knowledge~\cite{yu2023improving,peng2023check,huo2023retrieving}. Tools have a key role in computational verification. For example, CRITIC~\cite{goucritic} integrates calculators to validate the computational aspects of the model's output. Verifiers are often used for general reasoning tasks, which typically require additional data collection to train specialized models~\cite{cobbe2021training,li2022advance,lightman2023let}. However, this method relies on additional resources and computational costs, which may potentially limit its scalability.

\section{Self-Indicator}\label{section:theory}
Existing works~\cite{chen2024context,meng2022locating,geva2023dissecting} on self-verifying LLMs' outputs mainly focus on the case where the output is a single token. In this section, we propose Self-Indicator, which checks the correctness of the whole reasoning path and is capable of dealing with a much broader range of tasks.
Self-Indicator is based on the discovery that the correlations between correct reasoning paths are usually less complex than incorrect reasoning paths.
Specifically, we define the  correlation matrix $\mathbf{R}$ as:
\begin{equation}\label{relevance_matrix}
\mathbf{R} =
\begin{bmatrix}
\bm{h}_1^\top \bm{n}_1 & \bm{h}_1^\top \bm{n}_2 & \cdots & \bm{h}_1^\top \bm{n}_N \\
\bm{h}_2^\top \bm{n}_1 & \bm{h}_2^\top \bm{n}_2 & \cdots & \bm{h}_2^\top \bm{n}_N \\
\vdots & \vdots & \ddots & \vdots \\
\bm{h}_M^\top \bm{n}_1 & \bm{h}_M^\top \bm{n}_2 & \cdots & \bm{h}_M^\top \bm{n}_N
\end{bmatrix} \in \mathbb{R}^{M \times N}
\end{equation}
where $\{\bm{n}_1, \bm{n}_2,...,\bm{n}_N\}$ and $\{\bm{h}_1,\bm{h}_2,...,\bm{h}_M\}$ are the representations of tokens in problem $P$ and the reasoning path, respectively. Each $\bm{n}_i, \bm{h}_i\in \mathbb{R}^d$ is a $d$-dimensional vector. 

In this following parts, we will first motivate our choice of using the rank of the matrix as a proxy for correctness verification through theoretical analysis. Then, we describe our approach to leverage this metric for enhancing the reasoning performance for LLMs.

\subsection{Motivation}\label{section:motivation}
When solving a problem, a correct solution is expected to recognize and incorporate the key patterns inherent in the problem. In contrast, an incorrect solution may contain redundant or irrelevant information that does not align well with the problem's underlying structure. Based on this insight, we consider constructing the ``solution-problem'' correlation matrix $\mathbf{R}$ as shown in Figure~\ref{fig:preliminary}. Intuitively, for a correct solution, the correlation matrix primarily captures the alignment between meaningful patterns in the problem and their accurate reflection in the solution. As a result, this matrix tends to have a relatively low rank. On the other hand, incorrect solutions—due to their inclusion of spurious or inconsistent content—introduce additional variation, which can lead to an artificially increased rank in the correlation matrix. This contrast in matrix rank inspires us to use it as a promising signal for distinguishing correct and incorrect solutions. In the following, we will analyze this idea under a simplified idealized model to theoretically examine its feasibility and implications.

Formally, we consider a large language model (LLM) $\mathcal{M}$ as a function that takes a problem input $P=(\bm{n}_1, \bm{n}_2,...,\bm{n}_N)$ and outputs a representation $\mathcal{M}(\bm{n}_1, \ldots, \bm{n}_N)$ at the final position. To simplify the analysis, we focus on a single attention layer and adopt a setup following previous research~\cite{dai2023can,liu2024makes}:
\begin{equation}\label{single_attn}
  \mathcal{M}(\bm{n}_1, \ldots, \bm{n}_N) = \sum_{r=1}^{N} \frac{\bm{n}_N^\top \mathbf{W} \bm{n}_r}{\sum_{j} \bm{n}_N^\top \mathbf{W} \bm{n}_j} \cdot \bm{n}_r
\end{equation}
where $\mathbf{W}$ is the product of the key and query projection matrices, i.e., $\mathbf{W} \in \mathbb{R}^{d \times d} = (\mathbf{W}^k)^\top \mathbf{W}^q$. During the training process of $\mathcal{M}$, the commonly adopted objective is the cross-entropy loss based on token prediction. However, we notice that this loss can be interpreted as encouraging alignment between the model's output representations and the representations of the target tokens, since the token prediction probability is computed via the inner product of these two representations. 
Therefore, after training, in the optimal case, the predicted token representations of a correct solution should lie in the same direction of the representation of the golden-truth token $\bm{h}_i$.
Specifically, we have: 
\begin{equation}\label{condition1}
\bm{h}_1 \approx c_1\sum_{r=1}^{N} \frac{\bm{n}_N^\top \mathbf{W} \bm{n}_r}{\sum_{j} \bm{n}_N^\top \mathbf{W} \bm{n}_j} \cdot \bm{n}_r 
\end{equation}
\begin{equation}\label{condition}
\begin{aligned}
  \bm{h}_{i+1} \approx c_{i+1} & \sum_{r=1}^{N} \frac{\bm{h}_i^\top \mathbf{W} \bm{n}_r}{\sum_{j} \bm{h}_i^\top \mathbf{W} \bm{n}_j + \sum_{k=1}^{i} \bm{h}_i^\top \mathbf{W} \bm{h}_k} \cdot \bm{n}_r + \\
  &\left. \sum_{k=1}^{i} \frac{\bm{h}_i^\top \mathbf{W} \bm{h}_k}{\sum_{j} \bm{h}_i^\top \mathbf{W} \bm{n}_j + \sum_{k=1}^{i} \bm{h}_i^\top \mathbf{W} \bm{h}_k} \cdot \bm{h}_k\right)
\end{aligned}
\end{equation}
where $c_i\in \mathbb{R}$ is a scalar. This formulation captures how each solution token depends on an attention-weighted combination (defined by the parameter matrix $\mathbf{W}$) of both the original problem inputs and previously generated solution tokens. Denote $r = \text{rank}(\mathbf{W})$.
In practice, the parameter matrices of LLMs are often low-rank~\cite{oymak2019generalization,hsulanguage,saha2024compressing,halko2011finding}. Therefore, we assume that $r < min\{M, N,d\}$. 

For the correct solution, based on the analysis above, we have that each solution token representation $\bm{h}_i$ satisfies Eqs.~\eqref{condition1} and~\eqref{condition}. Without loss of generality, we further assume that $\{\bm{n}_1,...,\bm{n}_N\}$ are linearly independent (for repeated tokens, the columns of $\mathbf{R}$ would be identical, which does not affect the rank, so we disregard them). Under these setups, now we turn to estimate $\mathrm{rank}(\mathbf{R})$.

First, according to Lemma~\ref{thm:llema} in Appendix~\ref{append:lemma}, we prove that $\mathrm{rank}(\mathbf{R})$ is equal to 
\begin{equation}\label{final_rank}
\mathrm{rank}\left(
 \begin{bmatrix}
\bm{h}_1^\top \bm{n}_1 & \cdots & \bm{h}_1^\top \bm{n}_N \\
\bm{h}_1^\top \mathbf{W}_* \bm{n}_1 & \cdots & \bm{h}_1^\top \mathbf{W}_* \bm{n}_N \\
\bm{h}_1^\top \mathbf{W}^2_* \bm{n}_1 & \cdots & \bm{h}_1^\top \mathbf{W}^2_* \bm{n}_N \\
\vdots & \vdots & \vdots \\
\bm{h}_1^\top \mathbf{W}^{M-1}_*\bm{n}_1 & \cdots & \bm{h}_1^\top \mathbf{W}^{M-1}_*\bm{n}_N \\
\end{bmatrix}
\right)
\end{equation}
where $\mathbf{W}_* = \sum_{r=1}^{N}\mathbf{W} \bm{n}_r \cdot \bm{n}_r^\top$. Then, based on Eq.~\eqref{condition1}, $\bm{h}_1^\top \mathbf{W}^l_* \bm{n}_p=\sum_{r=1}^{N} \frac{\bm{n}_N^\top \mathbf{W} \bm{n}_r \cdot \bm{n}^\top_r \mathbf{W}^l_* \bm{n}_p}{\sum_{j} \bm{n}_N^\top \mathbf{W} \bm{n}_j} =\frac{\bm{n}_N^\top \mathbf{W}^{l+1}_* \bm{n}_p}{\sum_{j} \bm{n}_N^\top \mathbf{W} \bm{n}_j} $ for $\forall l\in \{1,...,M-1\}$. Therefore, Eq.~\eqref{final_rank} is further transformed into
\begin{equation}\label{final_rank1}
\mathrm{rank}\left(
 \begin{bmatrix}
\bm{n}_N^\top \mathbf{W}_* \bm{n}_1 & \cdots & \bm{n}_N^\top \mathbf{W}_* \bm{n}_N \\
\bm{n}_N^\top \mathbf{W}^{2}_* \bm{n}_1 & \cdots & \bm{n}_N^\top \mathbf{W}^{2}_* \bm{n}_N \\
\vdots & \vdots & \vdots \\
\bm{n}_N^\top \mathbf{W}^{M}_* \bm{n}_1 & \cdots & \bm{n}_N^\top \mathbf{W}^{M}_* \bm{n}_N \\
\end{bmatrix}
\right)=
\mathrm{rank}\left(
 \underbrace{\begin{bmatrix}
\bm{n}_N^\top \mathbf{W}_* \\
\bm{n}_N^\top \mathbf{W}^{2}_* \\
\vdots \\
\bm{n}_N^\top \mathbf{W}^{M}_* \\
\end{bmatrix} }_{M_1}\cdot 
\begin{bmatrix}
\bm{n}_1 & \cdots & \bm{n}_N \\
\end{bmatrix}
\right)
\end{equation}

Note that $\mathcal{A} \stackrel{\mathrm{def}}{=} \operatorname{\mathbf{span}}\{\bm{n}_N^\top \mathbf{W}_*, \bm{n}_N^\top \mathbf{W}^2_*, \dots, \bm{n}_N^\top \mathbf{W}^{M}_*\}$ is a Krylov subspace of the image space $\operatorname{Im}(\mathbf{W}_*)$. Denote $v=\mathrm{rank}(\mathbf{W}_*)  \leq r$, so the dimension of $\operatorname{Im}(\mathbf{W}_*)$ is also $v$. Let $\mathcal{S}=\{\bm{s}^\top_1, \dots, \bm{s}^\top_v\}$ be an orthogonal basis of $\operatorname{Im}(\mathbf{W}_*)$, where each $\bm{s}_i \in \mathbb{R}^d$. Then, it is easy to prove that the dimension of space $\mathcal{A}$ is strictly less than $v$ if and only if there exists a subset $\mathcal{S}^{\prime} \subseteq \mathcal{S}$ such that $\bm{n}_N$ lies in the kernel of the subspace spanned by $\mathcal{S}^{\prime}$, i.e., $\bm{n}_N \in \operatorname{\mathbf{ker}}\left(\operatorname{\mathbf{span}}(\mathcal{S})\right) := \mathcal{B}$. However, since the kernel space $\mathcal{B}$ is a strict linear subspace of $\mathbb{R}^d$, it has zero Lebesgue measure. Consequently, if $\bm{n}_N$ is sampled from a continuous distribution over $\mathbb{R}^d$, we have
\begin{equation}\label{leb_dim}
  \mathrm{P}(\dim(\mathcal{A}) = v) = 1.
\end{equation}

Eq.~\eqref{leb_dim} implies that if we express each $\bm{n}_N^\top \mathbf{W}^l_*= \sum_{j=1}^v a_{lj} \bm{s}_j^\top$ using the basis $\mathcal{S}$, then for
\begin{equation}
M_1 = 
\underbrace{\begin{pmatrix}
a_{11} & \dots & a_{1v} \\
\vdots & \ddots & \vdots \\
a_{M1} & \dots & a_{Mv}
\end{pmatrix}}_{M^{\prime}_1}
\begin{pmatrix}
\bm{s}_1^\top\\
\vdots \\
\bm{s}_v^\top
\end{pmatrix},
\end{equation}
we have $\mathrm{P}(\mathrm{rank}(M^{\prime}_1) = v) = 1$. Moreover, after rewriting Eq~\eqref{final_rank1} into
\begin{equation}\label{final_rank2}
\mathrm{rank}\left(M^{\prime}_1 \cdot 
\begin{pmatrix}
\bm{s}^\top_1 \\
\vdots \\
\bm{s}^\top_v
\end{pmatrix}
\cdot (\bm{n}_1, \ldots, \bm{n}_N)\right)=
\mathrm{rank}\left(M^{\prime}_1 \cdot 
\underbrace{\begin{pmatrix}
\bm{s}_1^\top \bm{n}_1 & \cdots & \bm{s}^\top_1 \bm{n}_N \\
\vdots & \ddots & \vdots \\
\bm{s}_v^\top \bm{n}_1 & \cdots & \bm{s}^\top_v \bm{n}_N
\end{pmatrix}}_{M_2}
\right),
\end{equation}
we note that if there exists an index set $K$ and coefficients $\{a_1, \ldots, a_k\}$ satisfying $\sum_{k \in K} a_k \bm{s}_k^\top \bm{n}_p = 0,\ \forall p\in\{1,...,N\}$, then we must have $\bm{n}_N \bot \sum_{k \in K} a_k \bm{s}_k$, i.e., $\bm{n}_N$ lies in the kernel space $\mathcal{B}$. Based on our previous analysis, we know that $\mathrm{P}(\bm{n}_N \in \mathcal{B}) = 0$, which implies that the probability that the rows of $M_2$ are linearly dependent is zero. Therefore, 
\begin{equation}
\mathrm{P}(\mathrm{rank}(M_2) = v) = 1.
\end{equation}

Finally, when $\mathrm{rank}(M^{\prime}_1) = v$ and $\mathrm{rank}(M_2)= v$, it follows that $\mathrm{rank}(M^{\prime}_1 M_2) = v$. As a result, we obtain that $\mathrm{P}(\mathrm{rank}(\mathbf{R}) = v) = 1$, where $v=\mathrm{rank}(\mathbf{W}_*)$ is a solution-independent value.  

In contrast, for an incorrect answer, denote its representations as $\{\bm{g}_1,...,\bm{g}_{M^\prime}\}$, where $M^\prime$ is the length of its token sequence. As an incorrect answer is typically not entirely wrong from start to end, we assume the existence of an index $\eta$ such that $\bm{g}_i=\bm{h}_i$ for $i \in \{1,...,\eta\}$, while for $i>\eta$, $\bm{g}_i$ contains noise or redundant information unrelated to the original problem. Similar to the analysis above, we know that the correlation matrix $\mathbf{R}^\prime$ formed by $\{\bm{g}_1,...,\bm{g}_\eta\}$ and $\{\bm{n}_1,...,\bm{n}_N\}$ has an expected rank of $v$. However, for $i>\eta$, the presence of irrelevant information will cause the correlation matrix formed by $\bm{g}_{\eta+1}$ to $\bm{g}_{M^\prime}$ has an expected rank of $min(N,M^\prime-\eta)$, and this matrix is linearly independent from $\mathbf{R}^\prime$. Therefore, the expected rank of the correlation matrix for the incorrect answer is $v+min(N,M^\prime-\eta)$, which is greater than the rank of the correct answer.

\textbf{To conclude, our analysis reveals that the cross-entropy loss of LLMs encourages a lower rank for the \emph{correlation matrix} of correct reasoning paths, while having less influence on incorrect reasoning paths.} As a result, the \emph{correlation matrix rank} is a suitable metric for evaluating the correctness of long reasoning paths generated by LLMs. 

It is important to emphasize that 1) \textbf{Applicability}: although our derivation is presented under the setup that both correct and incorrect answers are generated by the same inference LLM $\mathcal{M}$ (e.g., via multiple sampling), we argue that this restriction is not essential. This is because for a model that truly masters the reasoning, even if the surface form of the correct solution varies, its internal encoding should still minimize the cross-entropy loss and satisfy Eqs.~\eqref{condition1} and~\eqref{condition}, thus remaining compatible with our analysis. In Section~\ref{section:main_result_pre}, we will demonstrate that even when different LLMs are used for solution inference and for obtaining representations, our metric still exhibits strong effectiveness. 2) \textbf{Generalization Ability}: our analysis can also explain several empirical findings in existing literature. For instance, ~\cite{chen2024context} observed that a higher sharpness of context entropy often correlates with correct performance of LLMs on knowledge-based QA tasks. In our analysis, when a solution degenerates into a short few-token entity, a lower \emph{correlation matrix rank} typically indicates that the context activation is concentrated on a small subset of problem tokens, which leads to higher context entropy sharpness. Conversely, when information is more evenly spread across input tokens, the sharpness tends to be lower and the \emph{correlation matrix rank} will be correspondingly higher. Thus, our metric is a more general extension of entropy-based and attention-based analyses in prior work.

\subsection{Apply Self-Indicator to boost LLMs' reasoning performance}\label{section:our_method}
Based on our analyses, we now propose \textbf{Self-Indicator} that applies \emph{correlation matrix rank} to improve LLMs' reasoning performance. Specifically, given a problem $P$, we prompt an LLM to generate $K$ diverse solutions $\{S_1,...,S_K\}$ through multiple sampling. Each solution $S_k$ represents a complete reasoning path for solving $P$. Then, we compute the \emph{correlation matrix ranks} for each $S_k$ as follows. First, we feed $(P, S_k)$ into $\mathcal{M}$ with template: ``$\mathsf{Question: \{problem\}\ Answer: \{solution\}}$'' and compute the correlation matrix according to Eq.~\eqref{relevance_matrix}, denoted as $R^{QA}_{k}$. Here, we follow previous research~\cite{chen2024context,halawioverthinking} and use the hidden states of the 26-th transformer layer output as the representations\footnote{We also try using token embeddings, which results in a performance drop of 0.4\%–1.5\%.}. Then, we perform singular value decomposition (SVD): $R^{QA}_{k} = U_k \Sigma_k V_k^\top$. To avoid numerical noise, we apply a threshold $\delta$ and define the rank of $R^{QA}_k$ as the number of singular values in $\Sigma_k$ greater than $\delta$, which is further normalized by the number of tokens in the solution as the \emph{correlation matrix rank}. To ensure the robustness of representations, we repeat the above process using an inverse template: ``$\mathsf{Answer: \{solution\}\ Question: \{problem\}}$'' that swaps the order of the problem and solution (the necessity is discussed in Section~\ref{section:main_result_pre}). By doing so, we finally obtain two \emph{correlation matrix ranks} for each solution: $Rank^{QA}_{k}$ and $Rank^{AQ}_{k}$ and define the \textbf{Self-Indicator score} as\footnote{We experiment with different combinations of these ranks, but the results show no significant differences.}:
\begin{equation}
I_k = Rank^{QA}_k + Rank^{AQ}_k
\label{eq:indicator-add}
\end{equation}

This score serves as a proxy for the internal quality of each solution $S_k$, enabling us to prioritize more reliable reasoning paths. In practice, we rank all $K$ solutions based on their indicator scores and assign a weight to each solution using the formula $w_k = 1 + 0.5 \cdot (K - pos(k))$, where $pos(k)$ denotes the position of $S_k$ in the ascending order of indicator scores. Finally, we perform a weighted majority vote over all solutions using these weights to determine the final answer.

Our method has the following key advantages: 1) \textbf{Plug-and-Play}: Self-Indicator is entirely agnostic to the specific method used to generate multiple diverse solutions. This makes it applicable across a wide range of reasoning strategies and formats, such as Chain-of-Thought (CoT)~\cite{wei2022chain}. Besides, such a flexibility ensures that it can seamlessly integrate into other verification frameworks without significant adjustments (will be verified in Appendix~\ref{append_compatibility}). 2) \textbf{Low Complexity}: Self-Indicator scales efficiently with the number of samples $K$, with a time complexity of $\mathcal{O}(K)$. Moreover, it avoids the overhead of complex prompt engineering as required in previous works~\cite{weng2023large,ling2023deductive,miaoselfcheck}. 3) \textbf{Cost Efficient}: Unlike existing approaches that depend on expensive external verifiers (e.g., GPT-4) or require additional model training~\cite{goucritic, cobbe2021training, li2022advance, lightman2023let}, Self-Indicator is model-free and operates solely on the outputs generated by the LLM itself, requiring no additional model inference or training.
\section{Experiments}
In this section, we first empirically validate the effectiveness of \emph{correlation matrix rank} in distinguishing between correct and incorrect reasoning paths. Then, we demonstrate how our Self-Indicator leverages this property to enhance LLMs' reasoning performance.
\subsection{Validation of \emph{Correlation Matrix Rank}}\label{section_empirical}
\subsubsection{Experimental Setup}
We first conduct empirical studies to assess the effectiveness of \emph{correlation matrix rank} in distinguishing correct and incorrect solutions. We perform our experiments on MATH dataset~\cite{hendrycks2measuring}, which comprises over 12K problems from high school mathematics competitions. The solutions are typically long and structured, often requiring formal mathematical notation and deep problem-solving skills. Our goal is to investigate whether solutions with lower \emph{correlation matrix ranks} are more likely to be correct, thereby supporting the use of them in evaluating reasoning quality.

We begin by sampling the response of an LLM $\mathcal{M}$ for each problem in MATH and retain those problems that $\mathcal{M}$ answers incorrectly. Then, for each of these problems, we continue sampling twice (due to budget limit) until a correct solution is obtained. Problems that fail to achieve such requirements are excluded from further analysis. This process yields a curated dataset we refer to as \textbf{MATH-Pair}, which consists of original MATH problems $P$ paired with both a correct solution $S_{cor}$ and an incorrect solution $S_{inc}$ generated by $\mathcal{M}$.

\subsubsection{Main Results}\label{section:main_result_pre}
\textbf{Accuracy in Correctness Judgment.} On MATH-Pair, we calculate the Self-Indicator score $I_k$ in Section~\ref{section:our_method} and define the following decision rule: given two solutions, if solution 1 has lower $I_k$ than solution 2, we consider it to be the correct one. We validate this rule using LLaMA2–13B~\cite{touvron2023llama} 
\begin{wraptable}{r}{7.5cm}
    \small
    \centering
    \renewcommand{\arraystretch}{1.2} 
    \setlength{\tabcolsep}{1.5pt} 
    \caption{Decision Accuracy of our Self-Indicator score $I_k$ and $Rank^{QA}_k$ on MATH-Pair.}
    \label{pre_acc}
    \begin{tabular}{cccc}
    \toprule[1.5pt]
   Inference LLM & Representation LLM & $I_k$ & $Rank_k^{QA}$\\
   \hline LLaMA2–13B & LLaMA2–13B & 0.724 & 0.677 \\
          LLaMA3–70B & LLaMA3–70B & 0.783 & 0.755 \\
   \hline GPT-3.5-Turbo & LLaMA2–13B & 0.792 & 0.679 \\
   \bottomrule[1.5pt]
    \end{tabular}
\end{wraptable}
and LLaMA3–70B~\cite{meta2024introducing} as the backbone LLM $\mathcal{M}$ and perform a hyperparameter search over $\{0.75, 1.0, 1.25, 1.5, 1.75, 2.0\}$ to determine the best $\delta$. For comparison, we also implement this rule using the $Rank^{QA}_k$ alone. As shown in Table~\ref{pre_acc}, we observe that the decision accuracy is 72.4\% for LLaMA2-13B and 78.3\% for LLaMA3-70B, which directly validates that our Self-indicator score serves as a strong indicator for assessing the correctness of solutions. In addition, we observe that using $I_k$ achieves better decision accuracy compared to using $Rank^{QA}_k$ alone, indicating that integrating correlation matrix ranks from different templates is more robust and stable.

\textbf{Applicability to Closed-Source LLMs.} In practice, many powerful LLMs are closed-source, making it difficult to access their internal representations of problems and solutions. To examine whether it affects the applicability of our method, we further explore a cross-model setting: we use GPT-3.5-Turbo~\cite{chatgpt2023} as the inference model (i.e., to generate solutions), while employing LLaMA2-13B as the representation model (i.e., to compute the \emph{correlation matrix ranks}). 
\begin{wrapfigure}{r}{0.36\linewidth}
    \setlength{\abovecaptionskip}{0pt}
     \includegraphics[width=0.95\linewidth]{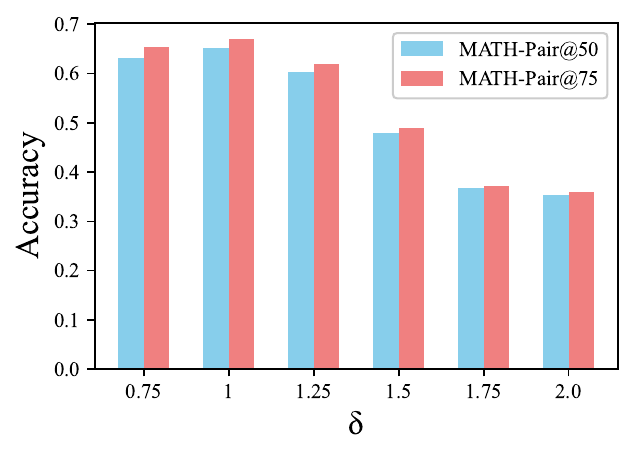}
    \caption{Decision Accuracy with length control at varying $\delta$. }
     \vspace{-0.2in}    \label{fig:multi_scale}
\end{wrapfigure}
As shown in the 3rd row of Table~\ref{pre_acc}, even when using different models for reasoning and representation, our proposed ranks still achieve strong accuracy in distinguishing between correct and incorrect solutions. This finding offers a practical insight: for closed-source LLMs, it is possible to leverage a smaller, open-source model to estimate the \emph{correlation matrix rank} of their outputs. This conclusion demonstrates the strong generality and flexibility of our metric.

\textbf{Impact of Solution Length.} In Eq.~\eqref{relevance_matrix}, it is easy to notice that the solution length $M$ may significantly influence the rank of correlation matrix. To control for this confounding factor, we construct two subsets of MATH-Pair: \textbf{MATH-Pair@50} and \textbf{MATH-Pair@75}, which contain samples where the length difference between the correct and incorrect solutions is less than 50 and 75 tokens, respectively. In Figure~\ref{fig:multi_scale}, we present the results across different $\delta$ using LLaMA2–13B as both the inference and representation model. We observe that when retains nearly all eigenvalues ($\delta\leq 1$), the accuracy still exceeds 65\%, which indicates that the discriminative power of our indicator score is not merely an artifact of solution length. In contrast, using a larger $\delta$ may filter out many informative eigenvalues of the correct answers, which potentially reverses the results.
\subsection{Effectiveness on Reasoning Performance}
\subsubsection{Experimental Setup}\label{section:experiment_setup}
\textbf{Datasets.} To evaluate the effectiveness of our proposed method, we conduct experiments on another two widely-used benchmark for mathematical reasoning: GSM8K~\cite{cobbe2021training} and AIME24~\cite{ai-mo2024aimo}, in addition to MATH~\cite{hendrycks2measuring}. GSM8K is a dataset of 8.5K high-quality grade school math word problems, covering arithmetic reasoning and requiring multi-step deduction. Each problem is annotated with a detailed step-by-step solution. AIME24 is a challenging benchmark composed of 30 problems from the American Invitational Mathematics Examination, focusing on advanced high school-level math. For all datasets, we adopt the public train-test split provided in their original releases for evaluation.

\textbf{Baselines.} For a fair comparison, we follow previous research~\cite{lightman2023let,ling2023deductive,miaoselfcheck} and compare our method (implementation details presented in Appendix~\ref{append:implementation}) with the following baselines that focus on solution verification \emph{without} solution refinement or improvement: 1) \emph{Original}: The inference LLM solves each problem with a single forward pass. 2) \emph{Self-Consistency} (SC)~\cite{wangself}: The model generates $K$ independent reasoning paths and selects the final answer via majority voting. 3) \emph{Self-Verification} (SV)~\cite{weng2023large}: As we found LLaMA-series struggles to generate high-quality rewritten candidate conclusions as required in the original paper, we implement the entire backward verification strategy using GPT-3.5-Turbo, where generated solutions are scored based on their consistency with the problem conditions. 4) \emph{Deductive-Verification} (DV)~\cite{ling2023deductive} and 5) \emph{SelfCheck} (SK)~\cite{miaoselfcheck}: Since LLaMA-family models may underperform on the complex prompt templates required by these methods, we directly quote the results on GPT-3.5 reported in their original papers.
\subsubsection{Main Results}\label{section:main_result}
\begin{table}[t]
\small
\centering
\setlength{\abovecaptionskip}{2pt}
\renewcommand{\arraystretch}{1.2} 
\setlength{\tabcolsep}{1.5pt} 
\caption{Answer Accuracy (\%) of different methods on GSM8K, MATH, and AIME24 datasets. The experiments are repeated five times, and our improvements are statistically significant with $p < 0.01$.}
\label{tab:accuracy-results}
\begin{tabular}{l|c|cccccccc|c}
\toprule[1.5pt]
\multirow{2}{*}{\textbf{Method}} & \multirow{2}{*}{\textbf{GSM8K}} & \multicolumn{8}{c|}{\textbf{MATH}} & \multirow{2}{*}{\textbf{AIME24}} \\
              \cline{3-10}         &           & \multicolumn{1}{l}{\footnotesize \textbf{Avg}} & \multicolumn{1}{l}{\footnotesize \textbf{Alg}} & \multicolumn{1}{l}{\footnotesize \textbf{Count}} & \multicolumn{1}{l}{\footnotesize \textbf{Geo}} & \multicolumn{1}{l}{\footnotesize \textbf{Itmd}} & \multicolumn{1}{l}{\footnotesize \textbf{Num}} & \multicolumn{1}{l}{\footnotesize \textbf{Pre-Alg}} & \multicolumn{1}{l|}{\footnotesize \textbf{Pre-Cal}} &  \\
\midrule
{\footnotesize LLaMA2-13B}        & \multicolumn{1}{l|}{44.6}    & \multicolumn{1}{l}{10.6} & \multicolumn{1}{l}{14.2} & \multicolumn{1}{l}{8.0} & \multicolumn{1}{l}{7.3} & \multicolumn{1}{l}{5.1} & \multicolumn{1}{l}{7.4} & \multicolumn{1}{l}{19.6} & \multicolumn{1}{l|}{5.9} & \multicolumn{1}{l}{3.3} \\
+ SC      & \multicolumn{1}{l|}{48.4}          & \multicolumn{1}{l}{15.1} & \multicolumn{1}{l}{19.3} & \multicolumn{1}{l}{10.1} & \multicolumn{1}{l}{10.9} & \multicolumn{1}{l}{9.5} & \multicolumn{1}{l}{14.2} & \multicolumn{1}{l}{22.4} & \multicolumn{1}{l|}{12.3} & \multicolumn{1}{l}{13.3} \\
+ SV       & \multicolumn{1}{l|}{45.7}       & \multicolumn{1}{l}{17.0} & \multicolumn{1}{l}{20.5} & \multicolumn{1}{l}{\textbf{15.1}} & \multicolumn{1}{l}{12.2} & \multicolumn{1}{l}{11.3} & \multicolumn{1}{l}{15.6} & \multicolumn{1}{l}{\textbf{23.7}} & \multicolumn{1}{l|}{16.0} & \multicolumn{1}{l}{3.3} \\
+ \textbf{Ours} & \multicolumn{1}{l|}{\textbf{49.3}~{\scriptsize\textcolor{ForestGreen}{↑4.7}}} & \multicolumn{1}{l}{\textbf{17.2}~{\scriptsize\textcolor{ForestGreen}{↑6.6}}} & \multicolumn{1}{l}{\textbf{20.9}~{\scriptsize\textcolor{ForestGreen}{↑6.7}}}& \multicolumn{1}{l}{10.3~{\scriptsize\textcolor{ForestGreen}{↑2.3}}} & \multicolumn{1}{l}{\textbf{13.2}~{\scriptsize\textcolor{ForestGreen}{↑5.9}}}& \multicolumn{1}{l}{\textbf{13.3}~{\scriptsize\textcolor{ForestGreen}{↑8.2}}} & \multicolumn{1}{l}{\textbf{17.0}~{\scriptsize\textcolor{ForestGreen}{↑9.6}}}& \multicolumn{1}{l}{22.8~{\scriptsize\textcolor{ForestGreen}{↑3.2}}} & \multicolumn{1}{l|}{\textbf{16.1}~{\scriptsize\textcolor{ForestGreen}{↑10.2}}} & \multicolumn{1}{l}{\textbf{16.7}~{\scriptsize\textcolor{ForestGreen}{↑13.4}}}\\
\midrule
{\footnotesize LLaMA3-70B}         & \multicolumn{1}{l|}{92.7}             & \multicolumn{1}{l}{49.1} & \multicolumn{1}{l}{65.1} & \multicolumn{1}{l}{46.0} & \multicolumn{1}{l}{35.5} & \multicolumn{1}{l}{28.7} & \multicolumn{1}{l}{40.0} & \multicolumn{1}{l}{71.1} & \multicolumn{1}{l|}{37.0} & \multicolumn{1}{l}{6.7}\\
+ SC            & \multicolumn{1}{l|}{96.7}   & \multicolumn{1}{l}{52.7} & \multicolumn{1}{l}{69.3} & \multicolumn{1}{l}{50.0} & \multicolumn{1}{l}{39.4} & \multicolumn{1}{l}{31.6} & \multicolumn{1}{l}{44.1} & \multicolumn{1}{l}{72.9} & \multicolumn{1}{l|}{41.7} & \multicolumn{1}{l}{20.0} \\
+ SV         & \multicolumn{1}{l|}{93.4}    & \multicolumn{1}{l}{46.9} & \multicolumn{1}{l}{61.5} & \multicolumn{1}{l}{45.8} & \multicolumn{1}{l}{34.2} & \multicolumn{1}{l}{27.0} & \multicolumn{1}{l}{37.2} & \multicolumn{1}{l}{64.6} & \multicolumn{1}{l|}{41.3} & \multicolumn{1}{l}{6.7} \\
+ \textbf{Ours}  & \multicolumn{1}{l|}{\textbf{96.8}~{\scriptsize\textcolor{ForestGreen}{↑4.1}}} & \multicolumn{1}{l}{\textbf{54.4}~{\scriptsize\textcolor{ForestGreen}{↑5.3}}} & \multicolumn{1}{l}{\textbf{70.4}~{\scriptsize\textcolor{ForestGreen}{↑5.3}}}& \multicolumn{1}{l}{\textbf{51.7}~{\scriptsize\textcolor{ForestGreen}{↑5.7}}} & \multicolumn{1}{l}{\textbf{41.8}~{\scriptsize\textcolor{ForestGreen}{↑6.3}}}& \multicolumn{1}{l}{\textbf{34.7}~{\scriptsize\textcolor{ForestGreen}{↑6.0}}} & \multicolumn{1}{l}{\textbf{45.6}~{\scriptsize\textcolor{ForestGreen}{↑5.6}}}& \multicolumn{1}{l}{\textbf{73.0}~{\scriptsize\textcolor{ForestGreen}{↑1.9}}} & \multicolumn{1}{l|}{\textbf{44.5}~{\scriptsize\textcolor{ForestGreen}{↑7.5}}} & \multicolumn{1}{l}{\textbf{23.3}~{\scriptsize\textcolor{ForestGreen}{↑16.6}}}\\
\midrule
{\footnotesize GPT-3.5-Turbo}         & \multicolumn{1}{l|}{83.8}          & \multicolumn{1}{l}{48.2} & \multicolumn{1}{l}{67.2} & \multicolumn{1}{l}{42.6} & \multicolumn{1}{l}{39.0} & \multicolumn{1}{l}{27.6} & \multicolumn{1}{l}{41.5} & \multicolumn{1}{l}{69.3} & \multicolumn{1}{l|}{27.3} & \multicolumn{1}{l}{10.0}\\
+ SC          & \multicolumn{1}{l|}{88.7}    & \multicolumn{1}{l}{55.2} & \multicolumn{1}{l}{77.2} & \multicolumn{1}{l}{48.9} & \multicolumn{1}{l}{42.5} & \multicolumn{1}{l}{35.5} & \multicolumn{1}{l}{45.1} & \multicolumn{1}{l}{74.3} & \multicolumn{1}{l|}{38.2}  & \multicolumn{1}{l}{13.3}\\
+ SV        & \multicolumn{1}{l|}{85.1}      & \multicolumn{1}{l}{50.7} & \multicolumn{1}{l}{69.0} & \multicolumn{1}{l}{48.2} & \multicolumn{1}{l}{36.2} &\multicolumn{1}{l}{31.1} & \multicolumn{1}{l}{45.9} & \multicolumn{1}{l}{67.4} & \multicolumn{1}{l|}{37.4} & \multicolumn{1}{l}{10.0} \\
+ DV        & \multicolumn{1}{l|}{87.1}      & \multicolumn{1}{l}{49.0\footnotemark}  & \multicolumn{1}{l}{/} &\multicolumn{1}{l}{/} & \multicolumn{1}{l}{/} & \multicolumn{1}{l}{/} & \multicolumn{1}{l}{/} & \multicolumn{1}{l}{/} & \multicolumn{1}{l|}{/} & \multicolumn{1}{l}{/}\\
+ SK         & \multicolumn{1}{l|}{88.1}     & \multicolumn{1}{l}{51.3}  & \multicolumn{1}{l}{/} &\multicolumn{1}{l}{/} & \multicolumn{1}{l}{/} & \multicolumn{1}{l}{/} & \multicolumn{1}{l}{/} & \multicolumn{1}{l}{/} & \multicolumn{1}{l|}{/} & \multicolumn{1}{l}{/} \\
+ \textbf{Ours} & \multicolumn{1}{l|}{\textbf{89.2}~{\scriptsize\textcolor{ForestGreen}{↑5.4}}}& \multicolumn{1}{l}{\textbf{56.9}~{\scriptsize\textcolor{ForestGreen}{↑8.7}}} & \multicolumn{1}{l}{\textbf{77.9}~{\scriptsize\textcolor{ForestGreen}{↑10.7}}}& \multicolumn{1}{l}{\textbf{51.1}~{\scriptsize\textcolor{ForestGreen}{↑8.5}}} & \multicolumn{1}{l}{\textbf{43.2}~{\scriptsize\textcolor{ForestGreen}{↑4.2}}}& \multicolumn{1}{l}{\textbf{38.4}~{\scriptsize\textcolor{ForestGreen}{↑10.8}}} & \multicolumn{1}{l}{\textbf{47.9}~{\scriptsize\textcolor{ForestGreen}{↑6.4}}}& \multicolumn{1}{l}{\textbf{75.3}~{\scriptsize\textcolor{ForestGreen}{↑6.0}}} & \multicolumn{1}{l|}{39.9~{\scriptsize\textcolor{ForestGreen}{↑12.6}}} & \multicolumn{1}{l}{\textbf{16.7}~{\scriptsize\textcolor{ForestGreen}{↑6.7}}}\\
\bottomrule[1.5pt]
\end{tabular}
\vspace{-10pt}
\end{table}

Table~\ref{tab:accuracy-results} presents the results of different methods across three backbones. First, our Self-Indicator consistently outperforms all baselines on both datasets. This demonstrates the effectiveness of our \emph{correlation matrix rank} metric in ranking and selecting high-quality reasoning paths. Second, on larger datasets, as the scale of the backbone model increases, the performance gains from our method become more pronounced (e.g., over 8\% for GPT-3.5-Turbo on MATH). This suggests that larger models are better able to leverage our proposed metric. Third, according to Appendix~\ref{append:variant}, different implementations of our Self-Indicator score achieve comparable and consistently top performance across different settings. This indicates the robustness of the core principle behind our approach, reflecting its efficacy irrespective of specific implementation choices. Fourth, Appendix~\ref{append:run_time} presents the runtime for computing the \emph{correlation matrix ranks}. The results show that our method introduces no significant inference-time overhead for LLMs, underscoring its cost-efficiency in practice.\footnotetext{This results is calculated based on the accuracy improvement in the original paper.}
\subsubsection{Analysis of Threshold $\delta$} 
To better understand the performance of our proposed Self-Indicator, we first investigate how the threshold $\delta$ affects its performance. As shown in Figure~\ref{fig:delta} and Figure~\ref{fig:delta_gsm} in Appendix~\ref{append:gsm8k}, we observe that regardless of the choice of $\delta$, our method consistently outperforms the best baseline Self-Consistency, which demonstrates the superb robustness and stability of our method. Second, for LLaMA2-13B, by comparing with Figure~\ref{fig:multi_scale}, we find a clear positive correlation between the reasoning accuracy and the correctness of solution assessment. This further demonstrates the effectiveness of our Self-Indicator score in evaluating solution quality and contribution to improving reasoning performance. Third, LLaMA2's accuracy decreases steadily as $\delta$ increases, whereas LLaMA3 and GPT-3.5 show more fluctuation rather than consistent degradation. We attribute these differences to variations in model scale and internal parameter distributions. This insight suggests that a fixed threshold may not be optimal across models, motivating future work on adaptive strategies to automatically select $\delta$ based on model-specific characteristics.

\subsubsection{Influence of the Number of Samples $K$} 
\begin{figure}[t]
  \centering
  \setlength{\abovecaptionskip}{2pt}
  \begin{subfigure}[b]{0.32\textwidth}
    \includegraphics[width=\textwidth]{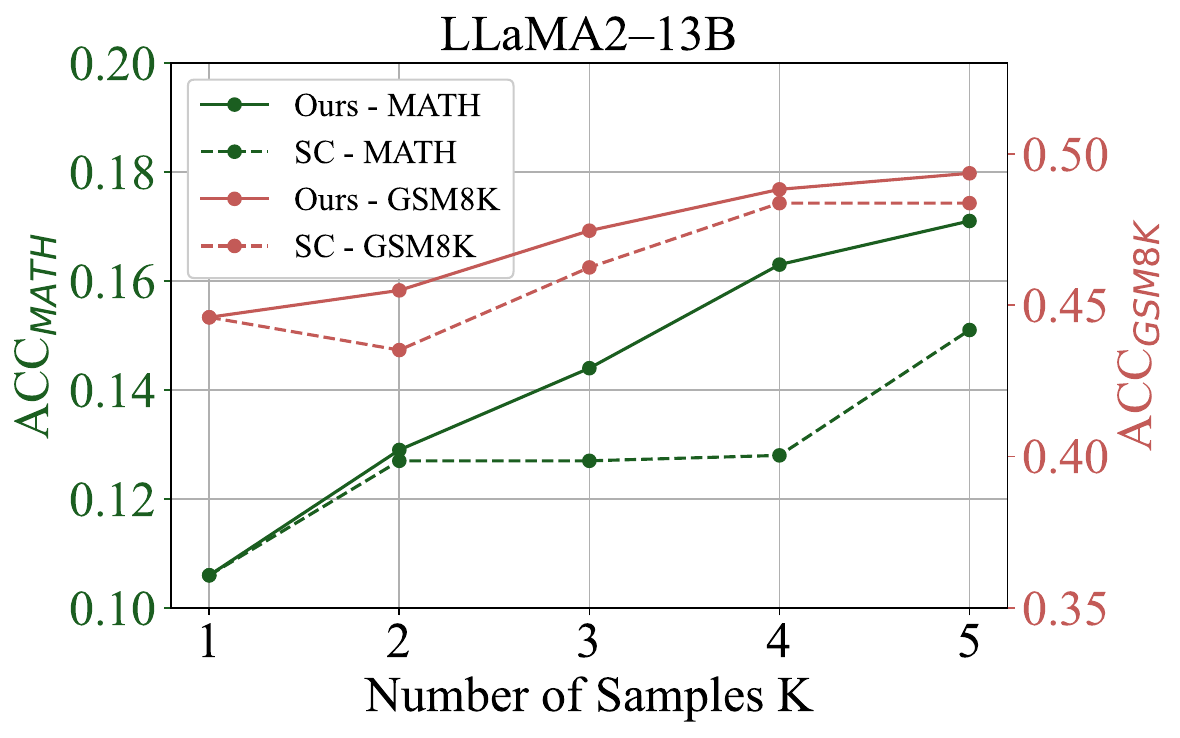}
  \end{subfigure}
  \begin{subfigure}[b]{0.32\textwidth}
    \includegraphics[width=\textwidth]{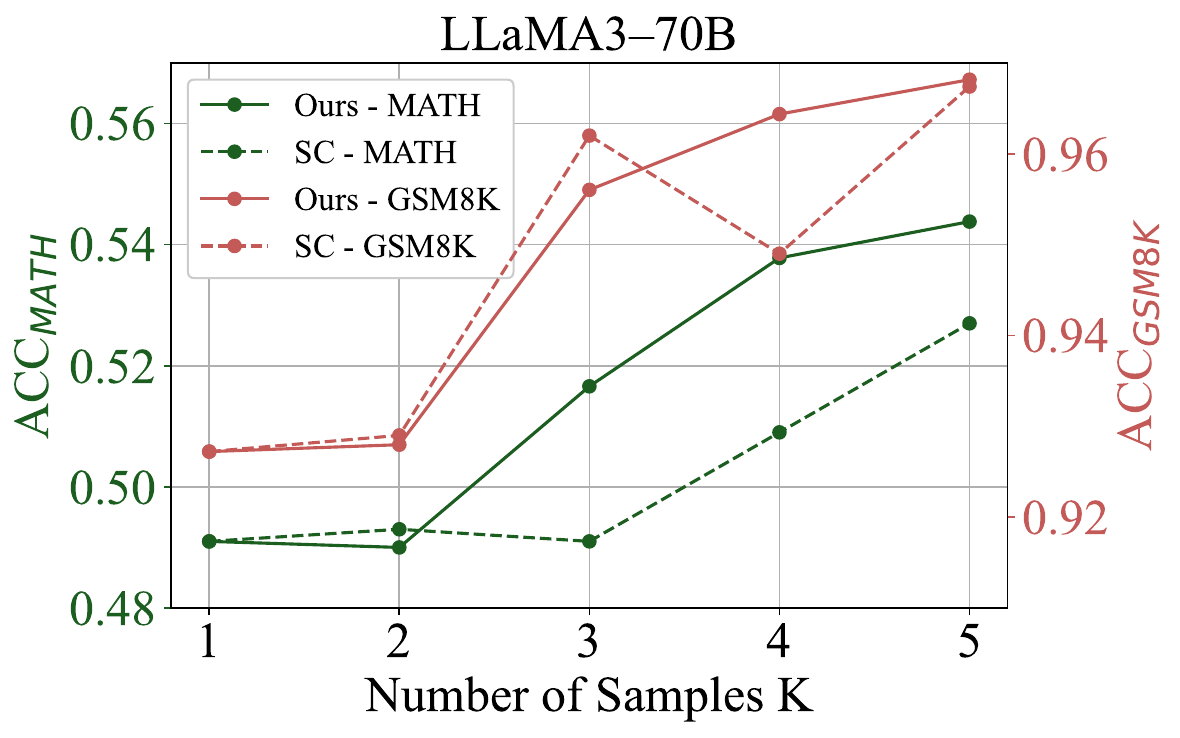}
  \end{subfigure}
  \begin{subfigure}[b]{0.32\textwidth}
    \includegraphics[width=\textwidth]{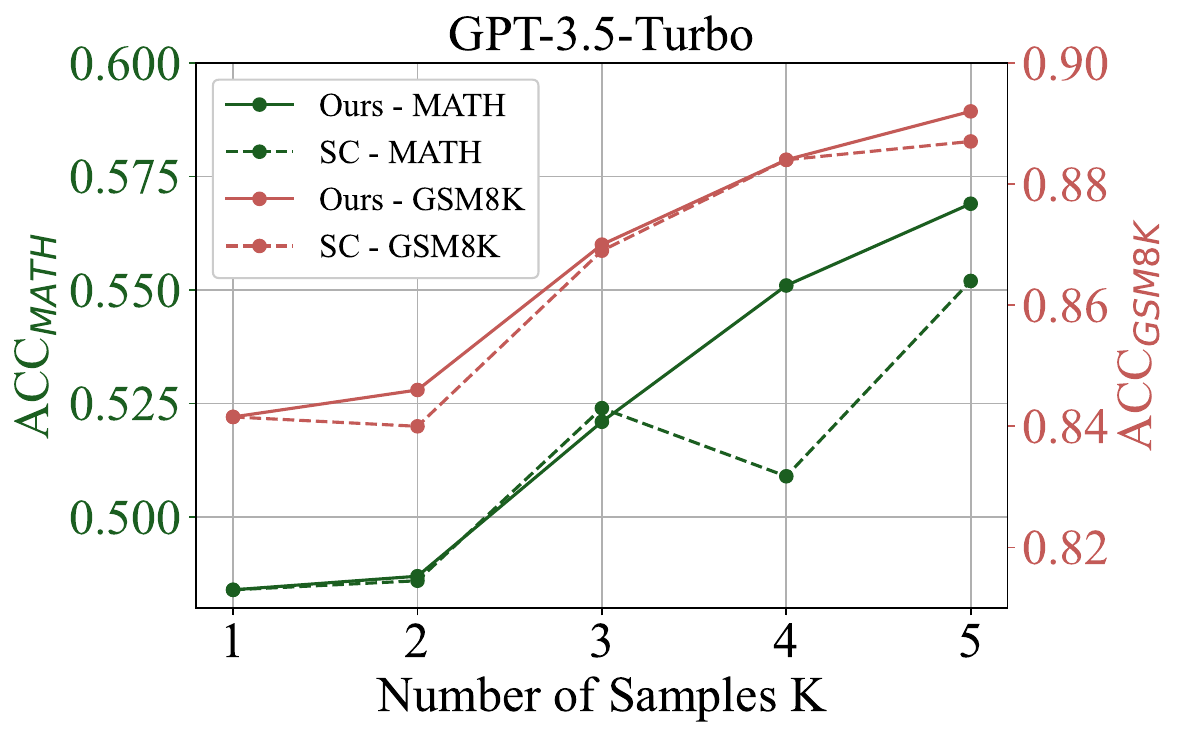}
  \end{subfigure}
  \caption{Performances with different $K$.}
  \label{fig:k}
\vspace{-15pt}
\end{figure}

We then analyze how the number of reasoning samples $K$ affects the effectiveness. As illustrated in Figure~\ref{fig:k}, increasing $K$ generally improves accuracy, as a larger number of samples provides more opportunities to produce correct answers. Moreover, our method still surpasses Self-Consistency for all $K\in\{1,2,3,4,5\}$. Notably, the advantage becomes more pronounced as $K$ increases. We think this is because, given more samples, our method can effectively identify more correct solutions and assign them greater weight in voting, whereas Self-Consistency treats all samples equally. This result highlights the strong discriminative ability of our proposed Self-Indicator score. Third, compared to GSM8K, our method achieves a larger improvement on the more challenging MATH dataset, suggesting that the benefits of our method are even more substantial when dealing with harder problems that require more sophisticated reasoning.
\begin{figure}[t]
  \centering
  \setlength{\abovecaptionskip}{0pt}
  \hspace{-15pt}
  \begin{subfigure}[b]{0.3\textwidth}
    \centering
    \includegraphics[width=\textwidth]{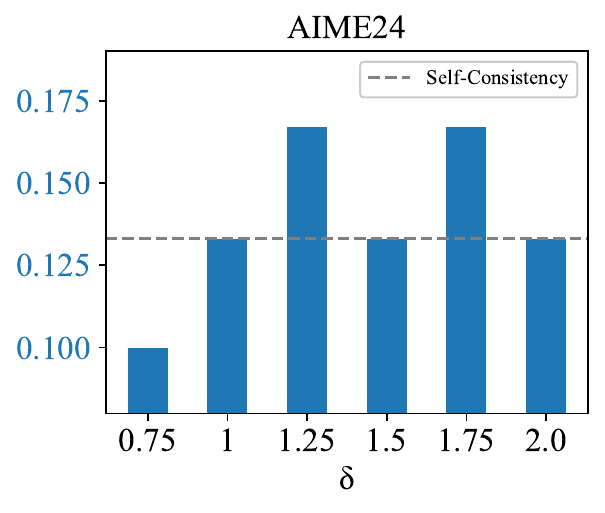}
  \end{subfigure}
  \hspace{5pt}
  \begin{subfigure}[b]{0.3\textwidth}
    \centering
    \includegraphics[width=\textwidth]{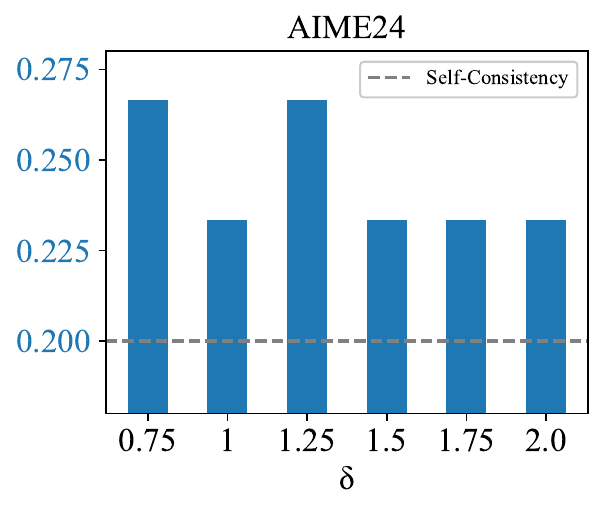}
  \end{subfigure}
   \hspace{5pt}
  \begin{subfigure}[b]{0.3\textwidth}
    \centering
    \includegraphics[width=\textwidth]{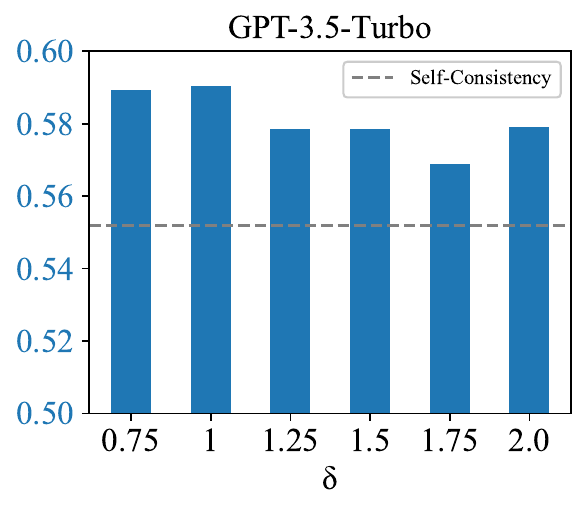}
  \end{subfigure}
  \caption{Performance with different $\delta$ on MATH dataset.}
  \label{fig:delta}
  \vspace{-10pt}
\end{figure}

\section{Conclusion} In this paper, we investigated whether LLMs inherently exhibit signals indicative of the correctness of their reasoning outputs. To this end, we proposed a novel metric \emph{correlation matrix rank} that captures internal consistency patterns within LLM-generated solutions and validated its effectiveness both theoretically and empirically. Building on this insight, we proposed the \emph{Self-Indicator} method, which leverages this rank to weight and select candidate reasoning paths. Our method is cost-efficient, requires no external knowledge or fine-tuning, and can be combined with existing verification techniques. Experimental results demonstrated its effectiveness, generality, and flexibility. In the future, we plan to explore its integration into broader reasoning frameworks and its extension to other NLP domains. Please refer to Appendix~\ref{Appendix_limit} for more discussions and future directions.

\begin{ack}
This work was supported by the National Natural science Foundation of China (Grant No.62477044, No.U23A20319), the Key Technologies R\&D Program of Anhui Province (No.202423k09020039), the Fundamental Research Funds for the Central Universities (No.WK2150110038). Zhenya Huang gratefully acknowledges the support of the Young Elite Scientists Sponsorship Program by CAST (No. 2024QNRC001).
\end{ack}
\bibliographystyle{plain}
\bibliography{bib}

\begin{thebibliography}{10}

\bibitem{ai-mo2024aimo}
{AI-MO}.
\newblock {AIMO Validation AMC Dataset}.
\newblock \url{https://huggingface.co/datasets/AI-MO/aimo-validation-amc}, 2024.
\newblock Accessed: 2025-03-29.

\bibitem{chang2024survey}
Yupeng Chang, Xu~Wang, Jindong Wang, Yuan Wu, Linyi Yang, Kaijie Zhu, Hao Chen, Xiaoyuan Yi, Cunxiang Wang, Yidong Wang, et~al.
\newblock A survey on evaluation of large language models.
\newblock {\em ACM transactions on intelligent systems and technology}, 15(3):1--45, 2024.

\bibitem{chen2024context}
Shiqi Chen, Miao Xiong, Junteng Liu, ZhengXuan Wu, Teng Xiao, Siyang Gao, and Junxian He.
\newblock In-context sharpness as alerts: an inner representation perspective for hallucination mitigation.
\newblock In {\em Proceedings of the 41st International Conference on Machine Learning}, pages 7553--7567, 2024.

\bibitem{cobbe2021training}
Karl Cobbe, Vineet Kosaraju, Mohammad Bavarian, Mark Chen, Heewoo Jun, Lukasz Kaiser, Matthias Plappert, Jerry Tworek, Jacob Hilton, Reiichiro Nakano, et~al.
\newblock Training verifiers to solve math word problems.
\newblock {\em arXiv preprint arXiv:2110.14168}, 2021.

\bibitem{dai2023can}
Damai Dai, Yutao Sun, Li~Dong, Yaru Hao, Shuming Ma, Zhifang Sui, and Furu Wei.
\newblock Why can gpt learn in-context? language models secretly perform gradient descent as meta-optimizers.
\newblock In {\em Findings of the Association for Computational Linguistics: ACL 2023}, pages 4005--4019, 2023.

\bibitem{geva2023dissecting}
Mor Geva, Jasmijn Bastings, Katja Filippova, and Amir Globerson.
\newblock Dissecting recall of factual associations in auto-regressive language models.
\newblock In {\em Proceedings of the 2023 Conference on Empirical Methods in Natural Language Processing}, pages 12216--12235, 2023.

\bibitem{geva2021transformer}
Mor Geva, Roei Schuster, Jonathan Berant, and Omer Levy.
\newblock Transformer feed-forward layers are key-value memories.
\newblock In {\em Proceedings of the 2021 Conference on Empirical Methods in Natural Language Processing}, pages 5484--5495, 2021.

\bibitem{goucritic}
Zhibin Gou, Zhihong Shao, Yeyun Gong, Yujiu Yang, Nan Duan, Weizhu Chen, et~al.
\newblock Critic: Large language models can self-correct with tool-interactive critiquing.
\newblock In {\em The Twelfth International Conference on Learning Representations}, 2024.

\bibitem{halawioverthinking}
Danny Halawi, Jean-Stanislas Denain, and Jacob Steinhardt.
\newblock Overthinking the truth: Understanding how language models process false demonstrations.
\newblock In {\em The Twelfth International Conference on Learning Representations}, 2024.

\bibitem{halko2011finding}
Nathan Halko, Per-Gunnar Martinsson, and Joel~A Tropp.
\newblock Finding structure with randomness: Probabilistic algorithms for constructing approximate matrix decompositions.
\newblock {\em SIAM review}, 53(2):217--288, 2011.

\bibitem{hendrycks2measuring}
Dan Hendrycks, Collin Burns, Saurav Kadavath, Akul Arora, Steven Basart, Eric Tang, Dawn Song, and Jacob Steinhardt.
\newblock Measuring mathematical problem solving with the math dataset.
\newblock In {\em Thirty-fifth Conference on Neural Information Processing Systems Datasets and Benchmarks Track (Round 2)}.

\bibitem{hsulanguage}
Yen-Chang Hsu, Ting Hua, Sungen Chang, Qian Lou, Yilin Shen, and Hongxia Jin.
\newblock Language model compression with weighted low-rank factorization.
\newblock In {\em International Conference on Learning Representations}, 2022.

\bibitem{huanglarge}
Jie Huang, Xinyun Chen, Swaroop Mishra, Huaixiu~Steven Zheng, Adams~Wei Yu, Xinying Song, and Denny Zhou.
\newblock Large language models cannot self-correct reasoning yet.
\newblock In {\em The Twelfth International Conference on Learning Representations}, 2024.

\bibitem{huo2023retrieving}
Siqing Huo, Negar Arabzadeh, and Charles Clarke.
\newblock Retrieving supporting evidence for generative question answering.
\newblock In {\em Proceedings of the annual international acm sigir conference on research and development in information retrieval in the Asia Pacific region}, pages 11--20, 2023.

\bibitem{li2022advance}
Yifei Li, Zeqi Lin, Shizhuo Zhang, Qiang Fu, Bei Chen, Jian-Guang Lou, and Weizhu Chen.
\newblock On the advance of making language models better reasoners.
\newblock {\em arXiv preprint arXiv:2206.02336}, 2, 2022.

\bibitem{lightman2023let}
Hunter Lightman, Vineet Kosaraju, Yuri Burda, Harrison Edwards, Bowen Baker, Teddy Lee, Jan Leike, John Schulman, Ilya Sutskever, and Karl Cobbe.
\newblock Let's verify step by step.
\newblock In {\em The Twelfth International Conference on Learning Representations}, 2024.

\bibitem{ling2023deductive}
Zhan Ling, Yunhao Fang, Xuanlin Li, Zhiao Huang, Mingu Lee, Roland Memisevic, and Hao Su.
\newblock Deductive verification of chain-of-thought reasoning.
\newblock {\em Advances in Neural Information Processing Systems}, 36:36407--36433, 2023.

\bibitem{liu2024makes}
Jiayu Liu, Zhenya Huang, Chaokun Wang, Xunpeng Huang, Chengxiang Zhai, and Enhong Chen.
\newblock What makes in-context learning effective for mathematical reasoning: A theoretical analysis.
\newblock {\em arXiv preprint arXiv:2412.12157}, 2024.

\bibitem{meng2022locating}
Kevin Meng, David Bau, Alex Andonian, and Yonatan Belinkov.
\newblock Locating and editing factual associations in gpt.
\newblock {\em Advances in neural information processing systems}, 35:17359--17372, 2022.

\bibitem{meta2024introducing}
AI~Meta.
\newblock Introducing meta llama 3: The most capable openly available llm to date.
\newblock {\em Meta AI}, 2024.

\bibitem{miaoselfcheck}
Ning Miao, Yee~Whye Teh, and Tom Rainforth.
\newblock Selfcheck: Using llms to zero-shot check their own step-by-step reasoning.
\newblock In {\em The Twelfth International Conference on Learning Representations}, 2024.

\bibitem{chatgpt2023}
OpenAI.
\newblock \url{https://chatgpt.com/}, 2023.

\bibitem{oymak2019generalization}
Samet Oymak, Zalan Fabian, Mingchen Li, and Mahdi Soltanolkotabi.
\newblock Generalization guarantees for neural networks via harnessing the low-rank structure of the jacobian.
\newblock {\em arXiv preprint arXiv:1906.05392}, 2019.

\bibitem{pan2023automatically}
Liangming Pan, Michael Saxon, Wenda Xu, Deepak Nathani, Xinyi Wang, and William~Yang Wang.
\newblock Automatically correcting large language models: Surveying the landscape of diverse self-correction strategies.
\newblock {\em arXiv preprint arXiv:2308.03188}, 2023.

\bibitem{peng2023check}
Baolin Peng, Michel Galley, Pengcheng He, Hao Cheng, Yujia Xie, Yu~Hu, Qiuyuan Huang, Lars Liden, Zhou Yu, Weizhu Chen, et~al.
\newblock Check your facts and try again: Improving large language models with external knowledge and automated feedback.
\newblock {\em arXiv preprint arXiv:2302.12813}, 2023.

\bibitem{petroni2019language}
Fabio Petroni, Tim Rockt{\"a}schel, Sebastian Riedel, Patrick Lewis, Anton Bakhtin, Yuxiang Wu, and Alexander Miller.
\newblock Language models as knowledge bases?
\newblock In {\em Proceedings of the 2019 Conference on Empirical Methods in Natural Language Processing and the 9th International Joint Conference on Natural Language Processing (EMNLP-IJCNLP)}, pages 2463--2473, 2019.

\bibitem{saha2024compressing}
Rajarshi Saha, Naomi Sagan, Varun Srivastava, Andrea Goldsmith, and Mert Pilanci.
\newblock Compressing large language models using low rank and low precision decomposition.
\newblock {\em Advances in Neural Information Processing Systems}, 37:88981--89018, 2024.

\bibitem{shinn2023reflexion}
Noah Shinn, Federico Cassano, Ashwin Gopinath, Karthik Narasimhan, and Shunyu Yao.
\newblock Reflexion: Language agents with verbal reinforcement learning.
\newblock {\em Advances in Neural Information Processing Systems}, 36:8634--8652, 2023.

\bibitem{touvron2023llama}
Hugo Touvron, Louis Martin, Kevin Stone, Peter Albert, Amjad Almahairi, et~al.
\newblock Llama 2: Open foundation and fine-tuned chat models.
\newblock {\em arXiv preprint arXiv:2307.09288}, 2023.

\bibitem{wangself}
Xuezhi Wang, Jason Wei, Dale Schuurmans, Quoc~V Le, Ed~H Chi, Sharan Narang, Aakanksha Chowdhery, and Denny Zhou.
\newblock Self-consistency improves chain of thought reasoning in language models.
\newblock In {\em The Eleventh International Conference on Learning Representations}, 2023.

\bibitem{wei2022chain}
Jason Wei, Xuezhi Wang, Dale Schuurmans, Maarten Bosma, Fei Xia, Ed~Chi, Quoc~V Le, Denny Zhou, et~al.
\newblock Chain-of-thought prompting elicits reasoning in large language models.
\newblock {\em Advances in neural information processing systems}, 35:24824--24837, 2022.

\bibitem{weng2023large}
Yixuan Weng, Minjun Zhu, Fei Xia, Bin Li, Shizhu He, Shengping Liu, Bin Sun, Kang Liu, and Jun Zhao.
\newblock Large language models are better reasoners with self-verification.
\newblock In {\em Findings of the Association for Computational Linguistics: EMNLP 2023}, pages 2550--2575, 2023.

\bibitem{xue2024decompose}
Shangzi Xue, Zhenya Huang, Jiayu Liu, Xin Lin, Yuting Ning, Binbin Jin, Xin Li, and Qi~Liu.
\newblock Decompose, analyze and rethink: Solving intricate problems with human-like reasoning cycle.
\newblock {\em Advances in Neural Information Processing Systems}, 37:357--385, 2024.

\bibitem{yu2023improving}
Wenhao Yu, Zhihan Zhang, Zhenwen Liang, Meng Jiang, and Ashish Sabharwal.
\newblock Improving language models via plug-and-play retrieval feedback.
\newblock {\em arXiv preprint arXiv:2305.14002}, 2023.

\bibitem{yuksekgonulattention}
Mert Yuksekgonul, Varun Chandrasekaran, Erik Jones, Suriya Gunasekar, Ranjita Naik, Hamid Palangi, Ece Kamar, and Besmira Nushi.
\newblock Attention satisfies: A constraint-satisfaction lens on factual errors of language models.
\newblock In {\em The Twelfth International Conference on Learning Representations}, 2024.

\end{thebibliography}
\appendix
\section{Problem Case}\label{append:pro_case}
The problem presented in Figure~\ref{fig:preliminary} is ``Sam decides to start a rumor. Sam tells the rumor to her three friends. Each of Sam's three friends then tells the rumor to three friends who have not heard the rumor. This continues for five total cycles. Sam telling her three friends was the first cycle. How many people, not including Sam, will have heard the rumor when the fifth cycle is complete?''. 

The ground-truth solution is ``At the end of one cycle, 3 people have heard the rumor. At the end of two cycles, \$3+9\$ people have heard the rumor. At the end of three cycles, \$3+9+27\$ people have heard the rumor, and so on. At the end of five cycles, \$3+9+27+81+243=\textbackslash boxed\{363\}\$ people have heard the rumor. \textbackslash n\textbackslash nNote: The formula \textbackslash[\textbackslash na+ar+ar\textasciicircum 2+\textbackslash cdots+ar\textasciicircum\{n-1\}=\textbackslash \textbackslash frac\{ar\textasciicircum\{n\}-a\}\{r-1\}\textbackslash n\textbackslash \textbackslash] for the sum of a geometric series may be used to sum \$3\textasciicircum 1+3\textasciicircum 2+\textbackslash cdots+3\textasciicircum 5\$''.
\section{Proof of Lemma 1}\label{append:lemma}
\begin{lemma}\label{thm:llema}
$\mathrm{rank}(\mathbf{R})$ is equal to the result in Eq.\eqref{final_rank}.
\end{lemma}

\emph{Proof}. According to Eq.~\eqref{condition}, for any $i\in\{1,...,M-1\}$ and $p\in\{2,...,N\}$, the inner product $\bm{h}_{i+1}^\top \bm{n}_p$ can be expanded as:

\begin{equation}\label{eq:delta_expansion} 
\bm{h}_{i+1}^\top \bm{n}_p = \sum_{r=1}^{N} \frac{\bm{h}_i^\top \mathbf{W} \bm{n}_r \cdot \bm{n}_r^\top \bm{n}_p}{\sum_{j} \bm{h}_i^\top \mathbf{W} \bm{n}_j + \sum_{k=1}^{i} \bm{h}_i^\top \mathbf{W} \bm{h}_k} + \underbrace{\sum_{k=1}^{i} \frac{\bm{h}_i^\top \mathbf{W} \bm{h}_k \cdot \bm{h}_k^\top \bm{n}_p}{\sum_{j} \bm{h}_i^\top \mathbf{W} \bm{n}_j + \sum_{k=1}^{i} \bm{h}_i^\top \mathbf{W} \bm{h}_k} }_{\Delta_1} 
\end{equation}
where $\Delta_1$ is indeed a linear combination of $\{\bm{h}_k^\top \bm{n}_p, k = 1, \ldots, i\}$. Therefore, 
\begin{equation}\label{linear_substitue}
 \mathrm{rank}(\mathbf{R}) = \mathrm{rank}\left(
 \begin{bmatrix}
\bm{h}_1^\top \bm{n}_1 & \cdots & \bm{h}_1^\top \bm{n}_N \\
\sum_{r=1}^{N} \bm{h}_1^\top \mathbf{W} \bm{n}_r\cdot \bm{n}_r^\top \bm{n}_1 & \cdots & \sum_{r=1}^{N} \bm{h}_1^\top \mathbf{W} \bm{n}_r \cdot \bm{n}_r^\top \bm{n}_N \\
\vdots & \vdots & \vdots \\
\sum_{r=1}^{N} \bm{h}_{M-1}^\top \mathbf{W} \bm{n}_r\cdot \bm{n}_r^\top \bm{n}_1 & \cdots & \sum_{r=1}^{N} \bm{h}_{M-1}^\top \mathbf{W} \bm{n}_r \cdot \bm{n}_r^\top \bm{n}_N \\
\end{bmatrix}
\right)
\end{equation}

Similar to Eq.~\eqref{eq:delta_expansion}, we can further derive that $\sum_{r=1}^{N} \bm{h}_{i+1}^\top \mathbf{W} \bm{n}_r\cdot \bm{n}_r^\top \bm{n}_p = $
\begin{equation}\label{eq:expansino2}
   \sum_{r=1}^{N} \sum_{t=1}^{N}\frac{\bm{h}_i^\top \mathbf{W} \bm{n}_r \cdot \bm{n}_r^\top \mathbf{W} \bm{n}_t \cdot \bm{n}_t^\top\bm{n}_p}{\sum_{j} \bm{h}_i^\top \mathbf{W} \bm{n}_j + \sum_{k=1}^{i} \bm{h}_i^\top \mathbf{W} \bm{h}_k} + \underbrace{\sum_{r=1}^{N} \sum_{k=1}^{i} \frac{\bm{h}_i^\top \mathbf{W} \bm{h}_k \cdot \bm{h}_k^\top \mathbf{W} \bm{n}_r \cdot \bm{n}_r^\top \bm{n}_p}{\sum_{j} \bm{h}_i^\top \mathbf{W} \bm{n}_j + \sum_{k=1}^{i} \bm{h}_i^\top \mathbf{W} \bm{h}_k}}_{\Delta_2}
\end{equation}
where $\Delta_2$ is also a linear combination of $\{\sum_{r=1}^{N} \bm{h}_k^\top \mathbf{W} \bm{n}_r\cdot \bm{n}_r^\top \bm{n}_p, k = 1, \ldots, i\}$. Denote $\mathbf{W}_* = \sum_{r=1}^{N}\mathbf{W} \bm{n}_r \cdot \bm{n}_r^\top$, Eq.~\eqref{linear_substitue} further equals:
\begin{equation}\label{linear_substitue2}
\mathrm{rank}(\mathbf{R}) = \mathrm{rank}\left(
 \begin{bmatrix}
\bm{h}_1^\top \bm{n}_1 & \cdots & \bm{h}_1^\top \bm{n}_N \\
\bm{h}_1^\top \mathbf{W}_* \bm{n}_1 & \cdots & \bm{h}_1^\top \mathbf{W}_* \bm{n}_N \\
\bm{h}_1^\top \mathbf{W}^2_* \bm{n}_1 & \cdots & \bm{h}_1^\top \mathbf{W}^2_* \bm{n}_N \\
\vdots & \vdots & \vdots \\
\bm{h}_{M-2}^\top \mathbf{W}^2_*\bm{n}_1 & \cdots & \bm{h}_{M-2}^\top \mathbf{W}^2_*\bm{n}_N \\
\end{bmatrix}
\right)
\end{equation}

Applying mathematical induction, we can finally derive that:
\begin{equation}
\mathrm{rank}(\mathbf{R}) = \mathrm{rank}\left(
 \begin{bmatrix}
\bm{h}_1^\top \bm{n}_1 & \cdots & \bm{h}_1^\top \bm{n}_N \\
\bm{h}_1^\top \mathbf{W}_* \bm{n}_1 & \cdots & \bm{h}_1^\top \mathbf{W}_* \bm{n}_N \\
\bm{h}_1^\top \mathbf{W}^2_* \bm{n}_1 & \cdots & \bm{h}_1^\top \mathbf{W}^2_* \bm{n}_N \\
\vdots & \vdots & \vdots \\
\bm{h}_1^\top \mathbf{W}^{M-1}_*\bm{n}_1 & \cdots & \bm{h}_1^\top \mathbf{W}^{M-1}_*\bm{n}_N \\
\end{bmatrix}
\right)
\end{equation}
\section{Implementation Details}\label{append:implementation}
To investigate the generality of our method across different model scales and families, we use three LLM backbones: LLaMA2–13B~\cite{touvron2023llama}, LLaMA3–70B~\cite{meta2024introducing}, and GPT-3.5-Turbo~\cite{chatgpt2023}. In all experiments, we set the number of reasoning samples $K=5$ and $\delta=1.75$ to avoid extensive hyperparameter selection. The temperature is fixed at 0.8 across all models and baselines to ensure a fair comparison. All experiments are conducted on a server with six NVIDIA A800 GPUs. Our codes are available at \url{https://github.com/Ljyustc/Self-Indicator}.
\section{More Experimental Results}
\subsection{Ablation on Different Self-Indicator Scores}\label{append:variant}
In Table~\ref{tab:accuracy-results-variant}, we present the performance of different strategies for combining $Rank_k^{QA}$ and $Rank_k^{AQ}$ to compute the Self-Indicator score. Self-Indicator$^{+}$ refers to the approach defined in Eq.~\eqref{eq:indicator-add} of Section~\ref{section:our_method}, while Self-Indicator$^{\times}$ denotes the variant that uses the product of $Rank_k^{QA}$ and $Rank_k^{AQ}$ as the Self-Indicator score.
\begin{table}[t]
\small
\centering
\setlength{\abovecaptionskip}{2pt}
\renewcommand{\arraystretch}{1.2} 
\setlength{\tabcolsep}{2pt} 
\caption{Answer Accuracy (\%) of two variants of our Self-Indicator.}
\label{tab:accuracy-results-variant}
\begin{tabular}{l|c|cccccccc|c}
\toprule[1.5pt]
\multirow{2}{*}{\textbf{Method}} & \multirow{2}{*}{\textbf{GSM8K}}  & \multicolumn{8}{c|}{\textbf{MATH}} & \multirow{2}{*}{\textbf{AIME24}}  \\
              \cline{3-10}          &          & \multicolumn{1}{l}{\footnotesize \textbf{Avg}} & \multicolumn{1}{l}{\footnotesize \textbf{Alg}} & \multicolumn{1}{l}{\footnotesize \textbf{Count}} & \multicolumn{1}{l}{\footnotesize \textbf{Geo}} & \multicolumn{1}{l}{\footnotesize \textbf{Itmd}} & \multicolumn{1}{l}{\footnotesize \textbf{Num}} & \multicolumn{1}{l}{\footnotesize \textbf{Pre-Alg}} & \multicolumn{1}{l|}{\footnotesize \textbf{Pre-Cal}} & \\
\midrule
LLaMA2–13B               & \multicolumn{1}{l|}{44.6}         & \multicolumn{1}{l}{10.6} & \multicolumn{1}{l}{14.2} & \multicolumn{1}{l}{8.0} & \multicolumn{1}{l}{7.3} & \multicolumn{1}{l}{5.1} & \multicolumn{1}{l}{7.4} & \multicolumn{1}{l}{19.6} & \multicolumn{1}{l|}{5.9} & \multicolumn{1}{l}{3.3} \\
+ Self-Indicator$^{+}$ & \multicolumn{1}{l|}{49.3~{\scriptsize\textcolor{ForestGreen}{↑4.7}}} & \multicolumn{1}{l}{\textbf{17.2}~{\scriptsize\textcolor{ForestGreen}{↑6.6}}} & \multicolumn{1}{l}{\textbf{20.9}~{\scriptsize\textcolor{ForestGreen}{↑6.7}}}& \multicolumn{1}{l}{10.3~{\scriptsize\textcolor{ForestGreen}{↑2.3}}} & \multicolumn{1}{l}{\textbf{13.2}~{\scriptsize\textcolor{ForestGreen}{↑5.9}}}& \multicolumn{1}{l}{\textbf{13.3}~{\scriptsize\textcolor{ForestGreen}{↑8.2}}} & \multicolumn{1}{l}{\textbf{17.0}~{\scriptsize\textcolor{ForestGreen}{↑9.6}}}& \multicolumn{1}{l}{22.8~{\scriptsize\textcolor{ForestGreen}{↑3.2}}} & \multicolumn{1}{l|}{16.1~{\scriptsize\textcolor{ForestGreen}{↑10.2}}}& \multicolumn{1}{l}{\textbf{16.7}~{\scriptsize\textcolor{ForestGreen}{↑13.4}}}\\
+ Self-Indicator$^{\times}$ & \multicolumn{1}{l|}{\textbf{49.5}~{\scriptsize\textcolor{ForestGreen}{↑4.9}}} & \multicolumn{1}{l}{17.0~{\scriptsize\textcolor{ForestGreen}{↑6.4}}} & \multicolumn{1}{l}{20.4~{\scriptsize\textcolor{ForestGreen}{↑6.2}}}& \multicolumn{1}{l}{10.5~{\scriptsize\textcolor{ForestGreen}{↑2.5}}} & \multicolumn{1}{l}{13.1~{\scriptsize\textcolor{ForestGreen}{↑5.8}}}& \multicolumn{1}{l}{12.7~{\scriptsize\textcolor{ForestGreen}{↑7.6}}} & \multicolumn{1}{l}{16.1~{\scriptsize\textcolor{ForestGreen}{↑8.7}}}& \multicolumn{1}{l}{23.3~{\scriptsize\textcolor{ForestGreen}{↑3.7}}} & \multicolumn{1}{l|}{\textbf{16.8}~{\scriptsize\textcolor{ForestGreen}{↑10.9}}}& \multicolumn{1}{l}{10.0~{\scriptsize\textcolor{ForestGreen}{↑6.7}}}\\
\midrule
LLaMA3–70B              & \multicolumn{1}{l|}{92.7}         & \multicolumn{1}{l}{49.1} & \multicolumn{1}{l}{65.1} & \multicolumn{1}{l}{46.0} & \multicolumn{1}{l}{35.5} & \multicolumn{1}{l}{28.7} & \multicolumn{1}{l}{40.0} & \multicolumn{1}{l}{71.1} & \multicolumn{1}{l|}{37.0} & \multicolumn{1}{l}{6.7}\\
+ Self-Indicator$^{+}$  & \multicolumn{1}{l|}{96.8~{\scriptsize\textcolor{ForestGreen}{↑4.1}}} & \multicolumn{1}{l}{\textbf{54.4}~{\scriptsize\textcolor{ForestGreen}{↑5.3}}} & \multicolumn{1}{l}{\textbf{70.4}~{\scriptsize\textcolor{ForestGreen}{↑5.3}}}& \multicolumn{1}{l}{51.7~{\scriptsize\textcolor{ForestGreen}{↑5.7}}} & \multicolumn{1}{l}{\textbf{41.8}~{\scriptsize\textcolor{ForestGreen}{↑6.3}}}& \multicolumn{1}{l}{\textbf{34.7}~{\scriptsize\textcolor{ForestGreen}{↑6.0}}} & \multicolumn{1}{l}{\textbf{45.6}~{\scriptsize\textcolor{ForestGreen}{↑5.6}}}& \multicolumn{1}{l}{\textbf{73.0}~{\scriptsize\textcolor{ForestGreen}{↑1.9}}} & \multicolumn{1}{l|}{\textbf{44.5}~{\scriptsize\textcolor{ForestGreen}{↑7.5}}}& \multicolumn{1}{l}{\textbf{23.3}~{\scriptsize\textcolor{ForestGreen}{↑16.6}}}\\
+ Self-Indicator$^{\times}$ & \multicolumn{1}{l|}{\textbf{97.1}~{\scriptsize\textcolor{ForestGreen}{↑4.4}}} & \multicolumn{1}{l}{54.2~{\scriptsize\textcolor{ForestGreen}{↑5.1}}} & \multicolumn{1}{l}{70.3~{\scriptsize\textcolor{ForestGreen}{↑5.2}}}& \multicolumn{1}{l}{\textbf{51.9}~{\scriptsize\textcolor{ForestGreen}{↑5.9}}} & \multicolumn{1}{l}{41.5~{\scriptsize\textcolor{ForestGreen}{↑6.0}}}& \multicolumn{1}{l}{34.4~{\scriptsize\textcolor{ForestGreen}{↑5.7}}} & \multicolumn{1}{l}{\textbf{45.6}~{\scriptsize\textcolor{ForestGreen}{↑5.6}}}& \multicolumn{1}{l}{72.9~{\scriptsize\textcolor{ForestGreen}{↑1.8}}} & \multicolumn{1}{l|}{43.9~{\scriptsize\textcolor{ForestGreen}{↑5.9}}}& \multicolumn{1}{l}{20.0~{\scriptsize\textcolor{ForestGreen}{↑13.3}}}\\
\midrule
GPT-3.5-Turbo         & \multicolumn{1}{l|}{83.8}            & \multicolumn{1}{l}{48.2} & \multicolumn{1}{l}{67.2} & \multicolumn{1}{l}{42.6} & \multicolumn{1}{l}{39.0} & \multicolumn{1}{l}{27.6} & \multicolumn{1}{l}{41.5} & \multicolumn{1}{l}{69.3} & \multicolumn{1}{l|}{27.3} & \multicolumn{1}{l}{10.0} \\
+ Self-Indicator$^{+}$  & \multicolumn{1}{l|}{\textbf{89.2}~{\scriptsize\textcolor{ForestGreen}{↑5.4}}}  & \multicolumn{1}{l}{\textbf{56.9}~{\scriptsize\textcolor{ForestGreen}{↑8.7}}} & \multicolumn{1}{l}{\textbf{77.9}~{\scriptsize\textcolor{ForestGreen}{↑10.7}}}& \multicolumn{1}{l}{\textbf{51.1}~{\scriptsize\textcolor{ForestGreen}{↑8.5}}} & \multicolumn{1}{l}{\textbf{43.2}~{\scriptsize\textcolor{ForestGreen}{↑4.2}}}& \multicolumn{1}{l}{\textbf{38.4}~{\scriptsize\textcolor{ForestGreen}{↑10.8}}} & \multicolumn{1}{l}{\textbf{47.9}~{\scriptsize\textcolor{ForestGreen}{↑6.4}}}& \multicolumn{1}{l}{75.3~{\scriptsize\textcolor{ForestGreen}{↑6.0}}} & \multicolumn{1}{l|}{39.9~{\scriptsize\textcolor{ForestGreen}{↑12.6}}}& \multicolumn{1}{l}{\textbf{16.7}~{\scriptsize\textcolor{ForestGreen}{↑6.7}}} \\
+ Self-Indicator$^{\times}$  & \multicolumn{1}{l|}{89.0~{\scriptsize\textcolor{ForestGreen}{↑5.2}}}  & \multicolumn{1}{l}{\textbf{56.9}~{\scriptsize\textcolor{ForestGreen}{↑8.7}}} & \multicolumn{1}{l}{\textbf{77.9}~{\scriptsize\textcolor{ForestGreen}{↑10.7}}}& \multicolumn{1}{l}{51.0~{\scriptsize\textcolor{ForestGreen}{↑8.4}}} & \multicolumn{1}{l}{43.0~{\scriptsize\textcolor{ForestGreen}{↑4.0}}}& \multicolumn{1}{l}{38.1~{\scriptsize\textcolor{ForestGreen}{↑10.5}}} & \multicolumn{1}{l}{47.7~{\scriptsize\textcolor{ForestGreen}{↑6.2}}}& \multicolumn{1}{l}{\textbf{75.6}~{\scriptsize\textcolor{ForestGreen}{↑6.3}}} & \multicolumn{1}{l|}{\textbf{40.3}~{\scriptsize\textcolor{ForestGreen}{↑13.0}}}& \multicolumn{1}{l}{10.0~{\scriptsize\textcolor{ForestGreen}{↑0.0}}} \\
\bottomrule[1.5pt]
\end{tabular}
\end{table}
\subsection{Runtime of Our Method}\label{append:run_time}
In Table~\ref{ana_runtime}, we report the actual runtime of our method for computing the correlation matrix ranks. For each input, the time required to compute $Rank^{QA}_{k}$ and $Rank^{AQ}_{k}$ scores remains under 0.5 seconds in all cases. Notably, even with large-scale models such as LLaMA3–70B and GPT-3.5-Turbo, the runtime remains efficient. These results demonstrate the cost-efficiency of our method, making it practical for real-world use without incurring significant inference-time latency.
\begin{table}
    \centering
    \renewcommand{\arraystretch}{1.} 
    \caption{Runtime(s) of our method for computing $Rank^{QA}_{k}$ and $Rank^{AQ}_{k}$ for each data.}
    \label{ana_runtime}
    \begin{tabular}{cccc}
    \toprule[1.5pt]
   Backbone & GSM8K & MATH & AIME24 \\
   \hline LLaMA2–13B & 0.241 & 0.228 & 0.483  \\
    LLaMA3–70B & 0.178 & 0.204 & 0.370 \\
    GPT-3.5-Turbo & 0.271 & 0.189 &  0.439 \\
   \bottomrule[1.5pt]
    \end{tabular}
\end{table}
\subsection{Influence of $\delta$ on GSM8K and AIME24 datasets}\label{append:gsm8k}
\begin{figure}[t]
  \centering
  \setlength{\abovecaptionskip}{2pt}
  \hspace{-15pt}
  \begin{subfigure}[b]{0.3\textwidth}
    \centering
    \includegraphics[width=\textwidth]{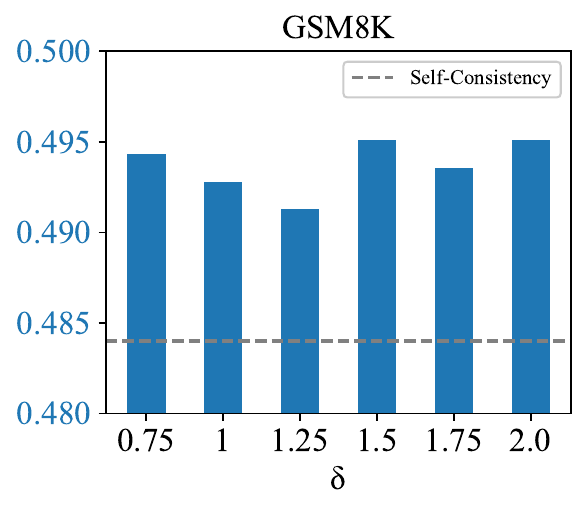}
    \includegraphics[width=\textwidth]{figures/delta-llama2-AIME-eps-converted-to.pdf}
    \caption{LLaMA2-13B}
  \end{subfigure}
  \hspace{5pt}
  \begin{subfigure}[b]{0.3\textwidth}
    \centering
    \includegraphics[width=\textwidth]{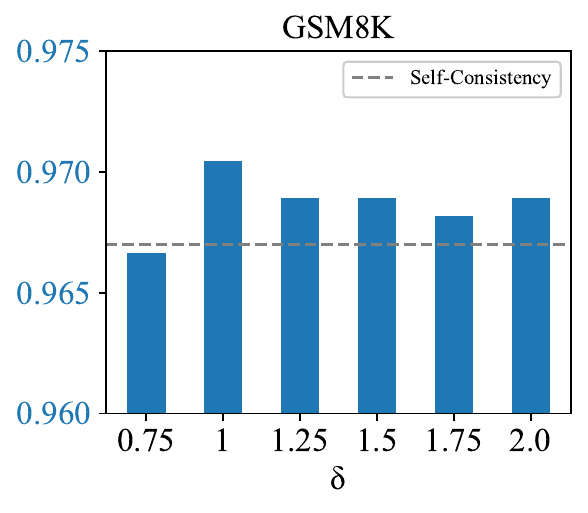}
    \includegraphics[width=\textwidth]{figures/delta-llama3-AIME-eps-converted-to.pdf}
   \caption{LLaMA3-70B}
  \end{subfigure}
   \hspace{5pt}
  \begin{subfigure}[b]{0.3\textwidth}
    \centering
    \includegraphics[width=\textwidth]{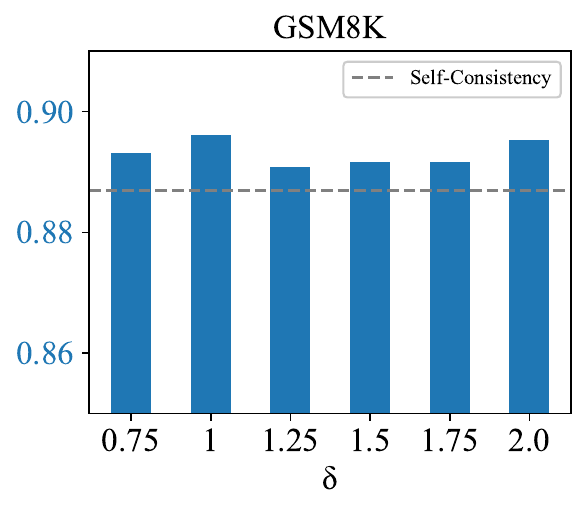}
    \includegraphics[width=\textwidth]{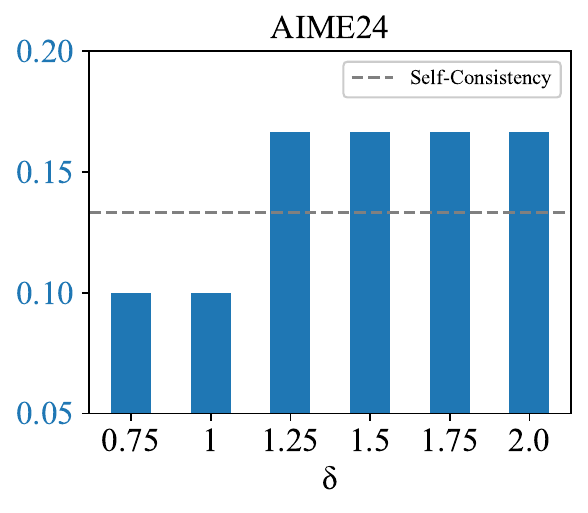}
   \caption{GPT-3.5-Turbo}
  \end{subfigure}
  \caption{Performance with different $\delta$ on GSM8K and AIME24 datasets.}
  \label{fig:delta_gsm}
\end{figure}

In addition to the results on the MATH dataset shown in Figure~\ref{fig:delta}, we present the performance of different $\delta$ on the GSM8K and AIME24 datasets in Figure~\ref{fig:delta_gsm}.
\subsection{Compatibility with Other Verification Methods}\label{append_compatibility} 
\begin{table}[t]
\small
\setlength{\abovecaptionskip}{2pt}
\centering
\renewcommand{\arraystretch}{1.1}
\caption{Performance of combining Self-Indicator and Self-Verification.}
\label{ana_compatibility}
\begin{tabular}{lccc}
\toprule[1.5pt]
Backbone & Method & $ACC_{\text{GSM8K}}$ & $ACC_{\text{MATH}}$ \\
\midrule
\multirow{3}{*}{LLaMA2–13B} 
    & Self-Verification &   45.7   & 17.0\\
    & Ours & 49.3 & 17.2 \\
    & Self-Verification+Ours & 48.9 & 18.0 \\
\midrule
\multirow{3}{*}{LLaMA3–70B} 
    & Self-Verification &   93.4  & 46.9\\
    & Ours & 96.8 & 54.5 \\
    & Self-Verification+Ours & 96.7 & 54.0 \\
\midrule
\multirow{3}{*}{GPT-3.5-Turbo} 
    & Self-Verification & 85.1  &   50.7   \\
    & Ours & 89.2 & 56.9 \\
    & Self-Verification+Ours & 88.9  & 57.1 \\
\bottomrule[1.5pt]
\end{tabular}
\end{table}
Here, we examine the complementary effect of Self-Indicator when integrated with other existing verification methods. Specifically, we sum our Self-Indicator scores with the scores computed by Self-Verification~\cite{weng2023large} for each solution, and then re-rank and vote based on the aggregated scores. In Table~\ref{ana_compatibility}, this fusion improves the accuracy of both LLaMA2-13B and GPT-3.5-Turbo on MATH dataset. This reflects the potential of our method to work in conjunction with other approaches. However, in other cases, the fused results fall short of the original Self-Indicator results reported in Table~\ref{tab:accuracy-results}. This is probably because there presents some noise in the self-verification scores (e.g., in Table~\ref{tab:accuracy-results}, Self-Verification performs even worse than Self-Consistency in these cases), as well as the suboptimal integration strategy when combining verification scores. This inspires us to explore how to better align with the strengths of different verification approaches in future research.
\section{Limitations and Future Work}\label{Appendix_limit}
First, our study primarily focuses on mathematical reasoning, which is a closed-book task, to validate the effectiveness of our method. In the future, we plan to explore the applicability of our theoretical framework and the \emph{Self-Indicator} method to broader tasks, such as multiple-choice questions and open-book scenarios. Second, in our theoretical analysis, we make certain simplifications regarding the training objectives and Transformer architecture. While these assumptions do not undermine the core advantages of our method, we aim to extend our analysis to more general and realistic settings in future work. Third, our experiments in Section~\ref{section_empirical} adopt a standard question-answering format to evaluate reasoning performance. For more sophisticated reasoning strategies, such as Tree-of-Thought or Program-of-Thought, the form of the reasoning process may significantly affect the assessment of correctness. This suggests that evaluating such strategies might require customized verification metrics beyond the current formulation. Finally, our method also shows promise in broader applications beyond evaluation. It can potentially serve as a data annotation tool to provide both correct and incorrect responses for training reward models, preference-based fine-tuning (e.g., DPO), and other alignment techniques. In addition, it can be incorporated into inference-time compute scaling approaches, such as Monte Carlo Tree Search (MCTS), to guide the selection or refinement of reasoning paths. We leave the exploration of these directions to future work.

\end{document}